%% file: 0_main.tex
\documentclass[letterpaper, 10 pt, journal, twocolumn]{Config/IEEEtran}
\usepackage[utf8]{inputenc}
\usepackage{bm}
\usepackage{amsmath}
\usepackage{amsthm}
\usepackage{amssymb}
\usepackage{mathtools}
\usepackage{graphicx}
\usepackage{subcaption}
\usepackage{stfloats}
\usepackage{xcolor}
\usepackage{multirow}

\input{Config/shortcutsRGR.tex}

\IEEEoverridecommandlockouts

\title{Task-space Synergies for Reaching using Upper-limb Prostheses}
\author{Ricardo Garcia-Rosas, Denny Oetomo, Chris Manzie, Ying Tan, and Peter Choong \thanks{This project is funded by the Valma Angliss Trust.} \thanks{R. Garcia-Rosas, D. Oetomo, C. Manzie, and Y. Tan are with the School of Electrical, Mechanical and Infrastructure Engineering, and P. Choong with the Department of Surgery, The University of Melbourne, VIC 3010, Australia. {\tt ricardog@student.unimelb.edu.au; \{doetomo,manziec,yingt,pchoong\}@unimelb.edu.au}.}}
\begin{document}

\maketitle

\begin{abstract}
Synergistic prostheses enable the coordinated movement of the human-prosthetic arm, as required by activities of daily living. This is achieved by coupling the motion of the prosthesis to the human command, such as the residual limb movement in motion-based interfaces. Previous studies demonstrated that developing human-prosthetic synergies in joint-space must consider individual motor behaviour and the intended task to be performed, requiring personalisation and task calibration. In this work, an alternative synergy-based strategy, utilising a synergistic relationship expressed in task-space, is proposed. This task-space synergy has the potential to replace the need for personalisation and task calibration with a model-based approach requiring knowledge of the individual user’s arm kinematics, the anticipated hand motion during the task and voluntary information from the prosthetic user. The proposed method is compared with surface electromyography-based and joint-space synergy-based prosthetic interfaces in a study of motor behaviour and task performance on able-bodied subjects using a VR-based transhumeral prosthesis. Experimental results showed that for a set of forward reaching tasks the proposed task-space synergy achieves comparable performance to joint-space synergies without the need to rely on time-consuming calibration processes or human motor learning. Case study results with an amputee subject motivate the further development of the proposed task-space synergy method.
\end{abstract}


\input{1_intro_vTS2.tex}
\input{3_synergy_vShort.tex}
\input{5_methodology_v3_1.tex}
\input{6_results_v3.tex}
\input{8_conclusion_v2.tex}

\section*{Acknowledgment}
\addcontentsline{toc}{section}{Acknowledgment}
The authors would like to thank Prof. Ian Gordon of the University of Melbourne Statistical Consulting Platform for his contribution to the statistical analysis of the results in this paper.

\bibliographystyle{ieeetr}  
\bibliography{library}  
\end{document}

%% file: Config/shortcutsRGR.tex
\newtheorem{assumption}{Assumption}

\newtheorem{remark}{Remark}

\newcommand{\frameOfRef}[1]{\prescript{#1}{}}
\newcommand{\transform}[2]{\prescript{#1}{#2}} 

\newcommand{\mbf}[1]{\mathbf{#1}}

\newcommand{\mbfd}[1]{\dot{\mathbf{#1}}}

\newcommand{\mcl}[1]{\mathcal{#1}}

\newcommand{\Real}{\mathbb{R}} 
 
\newcommand{\Natural}{\mathbb{N}} 

\newcommand{\xv}{\mbf{x}}

\newcommand{\zv}{\mbf{z}}

\newcommand{\qv}{\mbf{q}}
\newcommand{\qvd}{\mbfd{q}}

\newcommand{\pv}{\mbf{p}}

\newcommand{\vv}{\mbf{v}}

\newcommand{\rv}{\mbf{r}}

\newcommand{\Rv}{\mbf{R}}



%% file: 1_intro_vTS2.tex
%
%
\section{Introduction}

Synergy-based interfaces in prosthetics enable the coordinated motion of multiple degrees of freedom in the human-prosthetic arm as needed by many activities of daily living (ADLs) \cite{Merad2016}. The term ``synergy'' in the context of this work is used to refer to the coordination between multiple degrees of freedom in the human, and prosthetic, limb to achieve a task \cite{Latash2007, Santello2013}. Specifically, kinematic synergies relate the movement of the prosthetic device to the movement of the residual limb \cite{Kaliki2013}, e.g. regulating the movement of an elbow prosthesis as a function of the movement of the shoulder joint (residual limb) in a transhumeral amputation.  However, the kinematic synergies investigated in the literature to date have been found to be dependent on individual motor behaviour and the task, making them difficult to generalise and implement practically \cite{Merad2018, Legrand2018, Garcia-Rosas2018EMBC, Garcia-Rosas2019}.

Kinematic synergies are typically realised by a joint-space relationship, such that movements of the joints in the residual limb correspond to movements of the joints in the prosthesis \cite{Kaliki2013, Legrand2018, Garcia-Rosas2019, Akhtar2017, Blana2016, Merad2016, Alshammary2018}. So far, two approaches to realise joint-space kinematic synergies have been reported in the transhumeral prostheses literature: postural \cite{Kaliki2013, Akhtar2017} and differential synergies \cite{Legrand2018, Garcia-Rosas2019, Blana2016, Merad2016, Alshammary2018}.
Postural synergies define the human-prosthesis relationship in terms of joint displacements; i.e. for a given task and combination of joint displacements in the residual limb, there will be a resultant prosthesis joint displacement.
Differential synergies define the human-prosthesis relationship in terms of joint velocities. As such, at any given position, the movement of the prosthesis is a function of only the movement of the residual limb in both cases of kinematic synergies. Therefore, this relationship is dependent on the initial limb position and the task used to calibrate the synergy. Both types of synergies require extensive calibration, and have been found to be dependent on individual motor behaviour and the task to be performed \cite{Kaliki2013, Merad2018, Legrand2018, Garcia-Rosas2018EMBC, Akhtar2017}. While recent results have demonstrated that differential kinematic synergies can be personalised to their user during task execution \cite{Garcia-Rosas2019}, a library of task-specific and individually personalised synergies would be needed to provide the large range of movements required by ADLs \cite{Gates2015}.

The dependency of synergies on human individuality and the task performed is an unavoidable challenge to their application in fields such as prosthetics. Current methods do not utilise task information, such as the location of the target, as it is challenging to obtain it in a prosthetic setting. However, using this information may be beneficial in improving the robustness and performance of synergistic interfaces, and avert the need of generating a library of task specific synergies. While obtaining this information may be a challenge with current methods, this may be overcome with future technologies or interface modalities that assist with determining reaching intention and direction \cite{Li2017c}.

It is therefore necessary to develop and evaluate alternative strategies capable of explicitly using task information in the design of synergistic prosthetic interfaces. This serves to overcome task dependency and human motor behaviour variations while maintaining voluntary user control and involvement. In this work, an alternative synergy-based strategy, utilising a synergistic relationship expressed in task-space, is proposed. This method has the potential to replace the need for personalisation and task calibration with a model-based approach requiring knowledge of the individual user’s arm kinematics, the anticipated hand motion during the task and voluntary information from the prosthetic user. Using classic robotics tools, the proposed task-space synergy can be generated using the manipulator's Jacobian \cite{Whitney1969, Murray1994}, under the assumption that knowledge of the initial conditions and the desired end-effector (hand) motion is available. Here, a simplified case of such formulation under constrained motion is proposed. Location of the target is explicitly specified by the human user prior to the reaching motion, thus overcoming the assumption of the knowledge of the target location.

%
%
This paper presents the concept of task-space kinematic synergies in the context of forward reaching tasks with an elbow prosthesis. The performance of the resulting coordinated motion is compared against that produced in able-bodied motion, with a conventional sEMG interface, and a joint-space synergy \cite{Alshammary2018}. Experiments were carried out in a Virtual Reality Environment to allow the emulation of prosthesis use in able-bodied subjects. The outcomes show that, for a set of forward reaching tasks, the proposed task-space synergy achieved better performance than conventional sEMG and comparable performance to a joint-space synergy interface \cite{Alshammary2018} in terms of the closeness to able-bodied motor behaviour and task performance. The results achieved from a case study with an amputee subject motivate the further development of the proposed task-space synergy method.

%
%

%% file: 3_synergy_vShort.tex
%
%
\section{A Task-space Kinematic Synergy}
\begin{figure*}[ht]
\centering
\begin{subfigure}[t]{0.3\textwidth}
    \includegraphics[width=\textwidth]{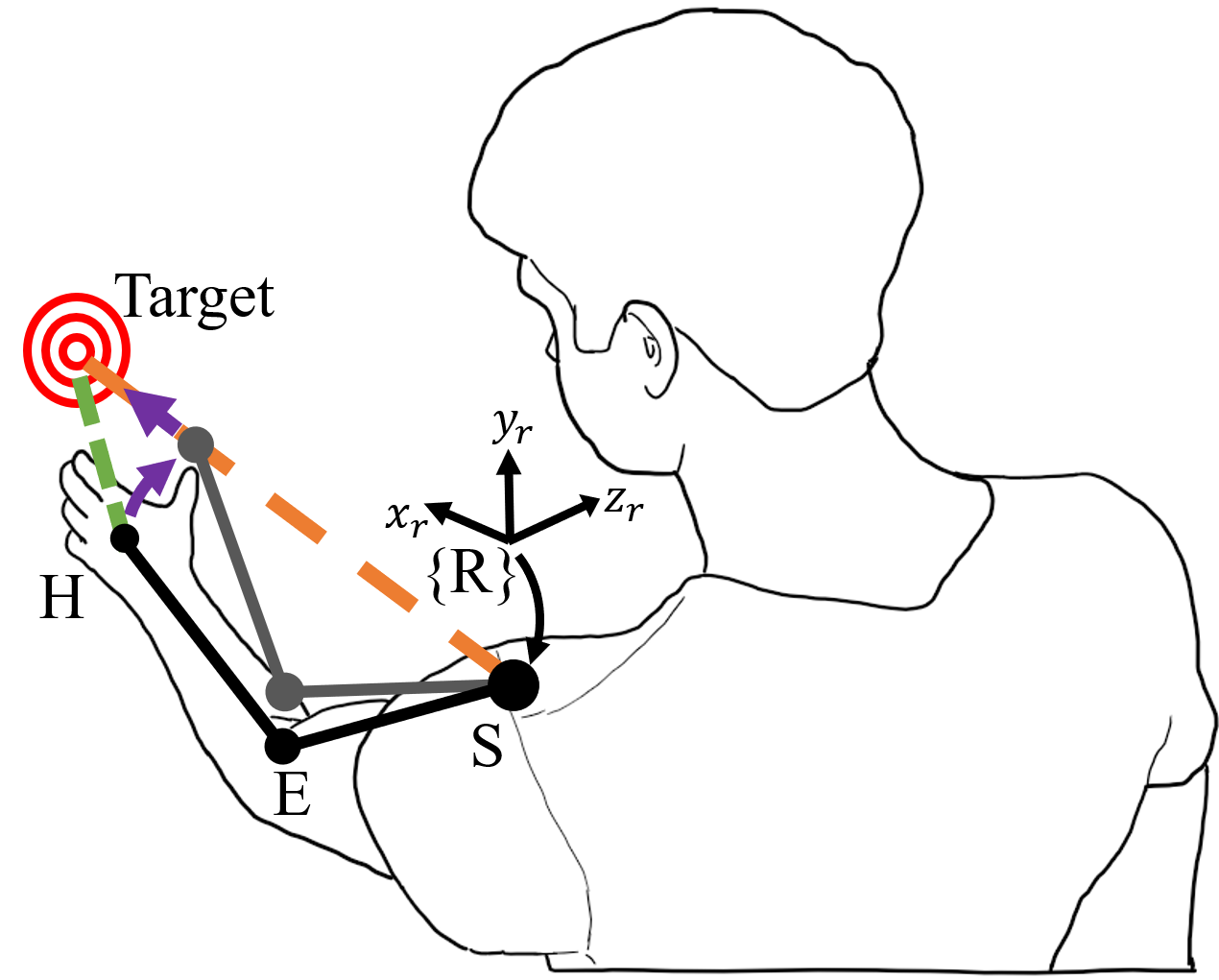}
    \caption{Target reaching example. The green dotted line shows the shortest path to target. The orange dotted line represents the forward straight reaching path. The reference frame $\{R\}$ has its origin at the shoulder joint and is attached to the shoulder.}
    \label{fig:reaching1}
\end{subfigure}
~ 
\begin{subfigure}[t]{0.3\textwidth}
    \includegraphics[width=\textwidth]{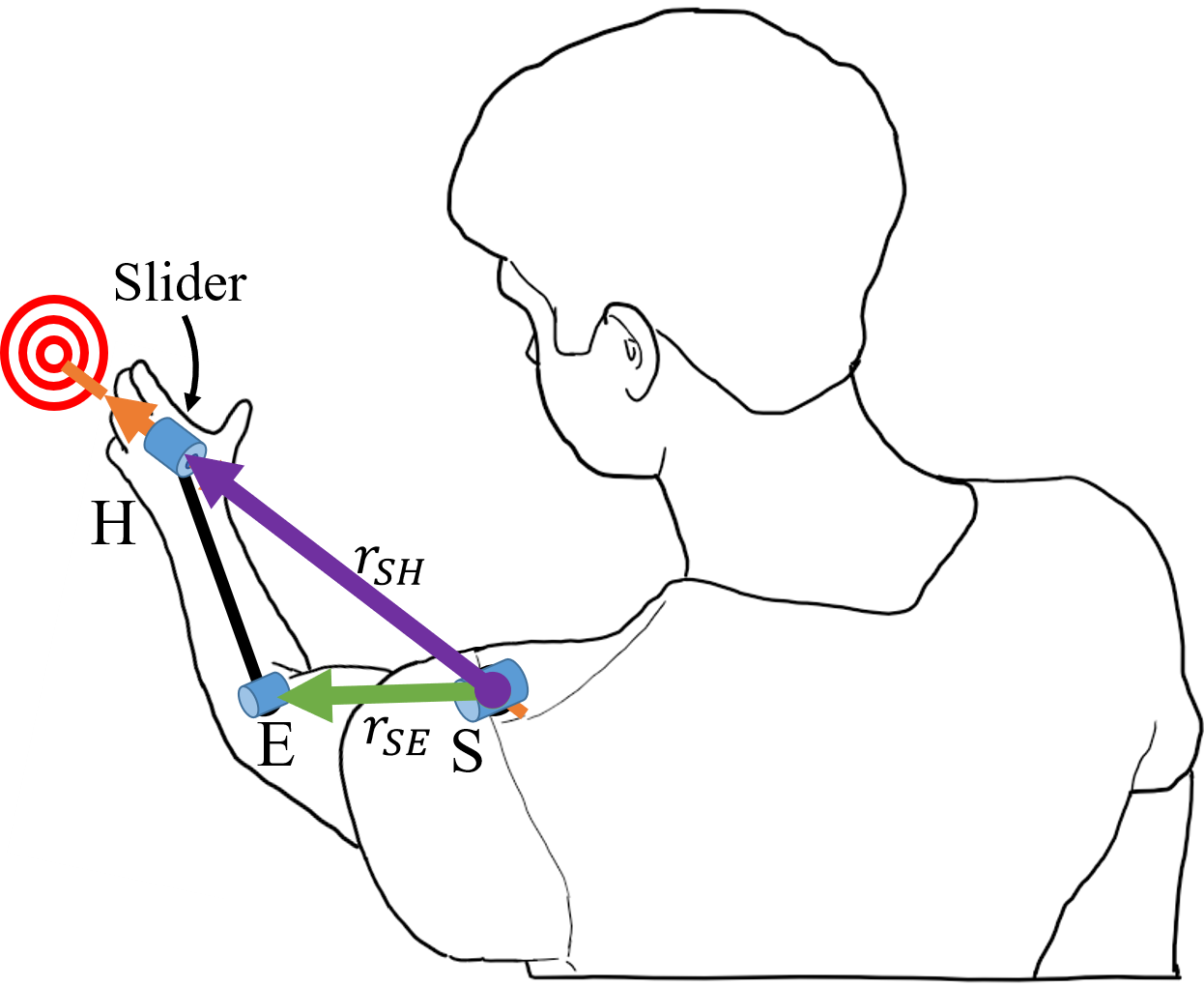}
    \caption{Simplified geometry for a reaching task by re-positioning the hand. The hand slides across the forward reaching path, defined by the vector ($r_{SH}$) given from the shoulder $\{S\}$ to the hand $\{H\}$ frame origins.}
    \label{fig:reaching2}
\end{subfigure}
~ 
\begin{subfigure}[t]{0.3\textwidth}
    \centering
    \includegraphics[width=0.7\textwidth]{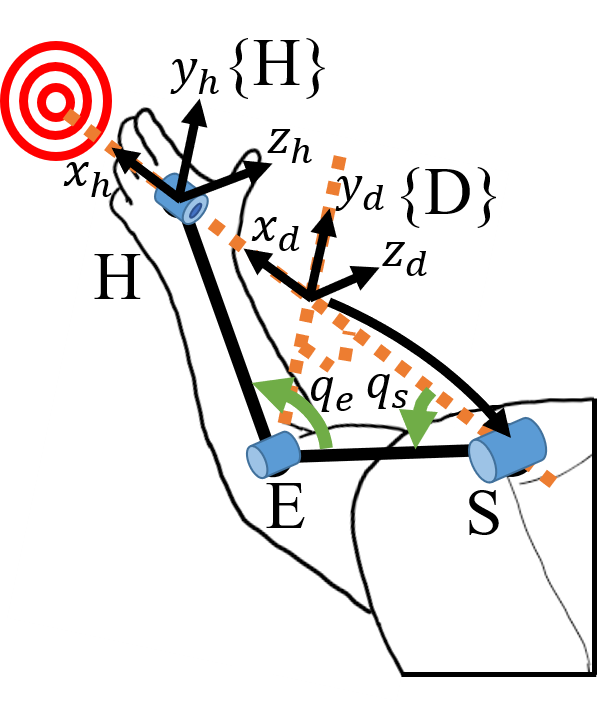}
    \caption{Definition of the direction of motion ($\{D\}$) and hand ($\{H\}$) frames that determine the forward reaching motion. The direction frame is defined by the triangle created by the shoulder (S), elbow (E), and hand (H) joints and it has its origin at the shoulder joint.}
    \label{fig:reaching3}
\end{subfigure}
\caption{Formulation of the forward reaching task and reference motion for the task-space synergy.}
\label{fig:reaching}
\end{figure*}

The objective of a task-space kinematic synergy is to determine the motion of the prosthesis given the motion of the residual limb and the desired hand path. However, knowing the exact desired hand path before execution of the task is a challenge as information on the desired target is only known to the prosthesis user. While there exist methods to obtain this information through other interfaces (e.g. gaze-tracking \cite{Li2017c}); here, a path-constraint approach is taken to allow the prosthetic user to convey the information of the target prior to commencing the reaching motion. This is done by the user aligning an imaginary straight line between the shoulder point, the target, and end effector (Figure \ref{fig:reaching1}) prior to a forward reaching motion. This determines the path of the hand reaching motion, described by the vector $r_{SH}$ (Figure \ref{fig:reaching2}). The path $r_{SH}$ can be thought of as the radial direction of a spherical coordinate centered at the shoulder point S. Once determined, the reaching motion, which is now constrained to move only along $r_{SH}$, can be produced through coupling the elbow prosthesis flexion-extension to the user's shoulder flexion-extension. This stage of the user aligning the hand along the target and the shoulder in a straight line will be referred to in this paper as the ``aiming'' stage, used to provide the information of the target in the proposed Task Space synergy approach.

The protocol for the user to determine the constrained path $r_{SH}$ is presented as follows. Let $\vv\in\Real^3$ be the hand velocity in task-space. $\qv_s$ and $\qvd_s$ are the residual limb joint angular displacement and velocity respectively (Figure \ref{fig:reaching3}). In the transhumeral amputation case under consideration, $\qv_s\in\Real^3$ and $\qvd_s\in\Real^3$. $\qv_e$ and $\qvd_e$ represent prosthesis joint angular displacement and velocity respectively (Figure \ref{fig:reaching3}), for the elbow prosthesis case $\qv_e\in\Real$ and $\qvd_e\in\Real$. 

The reference frame $\{R\}$ has its origin at the center of the shoulder joint and is attached to the shoulder with its $x$-axis pointing forward and the $y$-axis upwards w.r.t. the upper body. There are two reasons for the choice of the location of $\{R\}$. First, in practice, measurement of limb motion are obtained from IMUs and require a reference IMU to determine the limb's relative motion to the body. This IMU is typically placed on the upper-body. Second, to allow changes in the direction of motion with respect to the world coordinates. Since the motion of the arm is defined with respect to the trunk, any motion of the trunk (or rest of the body) allows the user to correct for initial errors in the direction of motion.

The coordinate frame $\{D\}$ has its origin at the shoulder joint and is attached to the direction of reach. The shoulder angle $\qv_s$ is given by the angle between the upper-arm (SE) and the direction of reach (SH). The frame $\{H\}$ has its origin at the centre of the hand (end-effector) with $x_h$ aligned to $x_d$.

Initially, the length of the upper arm ($d_{SE}$) and lower arm ($d_{EH}$) are measured and determined. Once the aiming process is completed, the following is carried out:
\renewcommand{\labelenumi}{\Alph{enumi}.}
\begin{enumerate}
    \item A coordinate frame $\{D\}$ is defined with its $x$-axis ($x_d$) aligned with the direction of reaching $r_{SH}$ and on the plane formed by the shoulder (S), elbow (E), and hand (H) joints.
    \item The frame transformation ($\transform{D}{R}\Rv$) is determined in order to express shoulder and elbow joint angular displacements with reference to frame $\{D\}$.
    \item The Jacobian matrix $\mcl{J}(\qv_s,\qv_e)$ is obtained to relate the incremental movement of the hand (end-effector) to the incremental joint displacement of the shoulder and elbow, taking into account the reaching path the hand motion is constrained to ($^{D}\dot{x}_h$).
    \item $\qvd_e$ is solved as a function of $\qv_s$, $\qv_e$, and $\qvd_s$ to determine the elbow motion reference that satisfies the desired task-space synergy.
\end{enumerate}
 These four steps are elaborated in the subsections below. The following assumption is used:
\begin{assumption}
    The angular position and velocities of the residual limb are measurable. 
\end{assumption}
\begin{remark}
    Inertial Measurement Unit (IMU)-based measurements of residual limb motion have been demonstrated for synergistic elbow prostheses in \cite{Alshammary2018, Merad2016a}.
\end{remark}
%
%
\subsection{Direction of Motion Reference Frame} 
The direction of motion frame $\{D\}$ determines the direction of the forward reaching task as exemplified in Figure \ref{fig:reaching}, and is defined w.r.t. the reference frame as follows: 

The $x_d$-axis is defined by the unit vector given in the direction from the shoulder joint center to the initial hand position $\frameOfRef{R}\pv$, as $\frameOfRef{R}\xv_d = \frac{\frameOfRef{R}\pv}{\|\frameOfRef{R}\pv\|_2}$. The $z_d$-axis is defined by the unit vector perpendicular to the plane given by the S-E-H triangle in the direction of $z_r$, as $\frameOfRef{R}\zv_d =  \frac{ \frameOfRef{R}\rv_{SE} \times \frameOfRef{R}\xv_d }{\|\frameOfRef{R}\rv_{SE} \times \frameOfRef{R}\xv_d\|_2}$, where $\frameOfRef{R}\rv_{SE}$ is the vector from the origin of the shoulder frame $\{S\}$ to the origin of the elbow frame $\{E\}$. The $y_d$-axis follows the right hand rule. 
    
With this the direction of motion reference frame $\{D\}$ is defined and motion of the arm and hand can be expressed with it as reference. This concludes step A. 

%
%
\subsection{Frame Transformation} 
Assuming that measured data is obtained in frame $\{R\}$ then it is necessary to transform these angles to the task frame $\{D\}$. This assumption arises from the common use of relative measurements from two IMUs to produce residual limb pose measurements \cite{Alshammary2018, Merad2016a}. This transformation can be performed through the Euler angles ($\alpha$, $\beta$, $\gamma$) in a Y-X-Z sequence. The transformation angles can be obtained as follows:

The first rotation ($\alpha$) is along the $y_r$-axis. $\alpha$ can be obtained from $\frameOfRef{R}\xv_d$, and is defined as $\alpha = \tan^{-1}{\left ( \frac{\frameOfRef{R}\xv_{d,x}}{\frameOfRef{R}\xv_{d,z}} \right )}$, where $\frameOfRef{R}\xv_{d,x}$ and $\frameOfRef{R}\xv_{d,z}$ are the $x$ and $z$ components of $\frameOfRef{R}\xv_{d}$. The second rotation ($\beta$) is along the $x_r$-axis. $\beta$ can be obtained from $\frameOfRef{R}\zv_d$, and is defined as $\beta = \tan^{-1}{\left ( \frac{\frameOfRef{R}\zv_{d,y}}{\frameOfRef{R}\zv_{d,z}} \right )}$. The last rotation ($\gamma$) is along the $z_r$-axis. $\gamma$ can be obtained from $\frameOfRef{R}\xv_d$, and is defined as $\gamma = \tan^{-1}{\left ( \frac{\frameOfRef{R}\xv_{d,y}}{\frameOfRef{R}\xv_{d,x}} \right )}$.

The transformation is therefore given by:
\begin{equation}
    \transform{D}{R}\Rv = \Rv_{y}(\alpha)\Rv_{x}(\beta)\Rv_{z}(\gamma),
\end{equation}
where $\Rv_{x}(\alpha)$ is an $x$-axis standard rotation of the given angle $\alpha$. Thus it is possible to transform arm motion data from the reference frame $\{R\}$ to the task frame $\{D\}$. This concludes step B.

%
%
\subsection{Arm Jacobian}
By performing the adequate transformations, as previously presented, the forward reaching task can be simplified to a 2D task on the $x-y$ plane of the $\{D\}$ frame. This simplifies the kinematics required for the synergy to those of a 2-DOF planar arm. The differential kinematics of the simplified arm are well known and given by \cite{Murray1994}:
\begin{equation}
\label{eq:simpDiffKine}
    \bar{\vv} = \begin{bmatrix} ^{D}\dot{x}_h \\ ^{D}\dot{y}_h \end{bmatrix} = \begin{bmatrix}
    -\ell_{u} S_s - \ell_{l} S_{se} & -\ell_{l} S_{se}\\
    \ell_{u} C_s + \ell_{l} C_{se} & \ell_{l} C_{se}
    \end{bmatrix}\begin{bmatrix} \dot{q}_s \\ \dot{q}_e \end{bmatrix}.
\end{equation}
where $C_s = \cos{(\bar{q}_{s})}$, $S_s = \sin{(\bar{q}_{s})}$, $C_{se} = \cos{(\bar{q}_{s} + \bar{q}_{e})}$, $S_{se} = \sin{(\bar{q}_{s} + \bar{q}_{e})}$, $\bar{q}_{s} = \frameOfRef{D}q_{s}$, and $\bar{q}_{e} = \frameOfRef{D}q_{e}$. $\ell_{u}$and $\ell_{l}$ are the lengths of the upper and lower arm respectively. This concludes step C.

%
%
\subsection{Elbow Motion Reference}
As previously discussed, through the simplification of the task only forward reaching motion is required along the direction of motion plane, i.e. only motion along the $x_d$-axis. This means that it is desired to satisfy the constraint $^{D}\dot{y}_h = 0$. With this constraint in consideration, it is possible to solve for the desired elbow velocity $\dot{q}_e$ from \eqref{eq:simpDiffKine}, such that
\begin{equation}
\label{eq:taskSpaceSynergy}
    \dot{q}_e = -\frac{\ell_{u} C_s + \ell_{l} C_{se}}{\ell_{l}C_{se}} \dot{q}_s.
\end{equation}
It is important to note that an elbow singularity occurs when $\bar{q}_{s} + \bar{q}_{e} = n\pi$, $n = 0, 2, 4, \cdots$ \cite{Oetomo2009}. This happens when the arm is at full extension or folded back into itself. The first configuration can be avoided by limiting the range of motion of the elbow such that $\| q_{e} \| > \epsilon_q$, where $\epsilon_q$ is a positive non-zero constant. The second configuration is not physically possible. This concludes step D.


%% file: 5_methodology_v3_1.tex
%
%
\section{Experimental Methodology}

\begin{figure*}[ht]
\centering
\begin{subfigure}[t]{0.29\textwidth}
    \includegraphics[width=\textwidth]{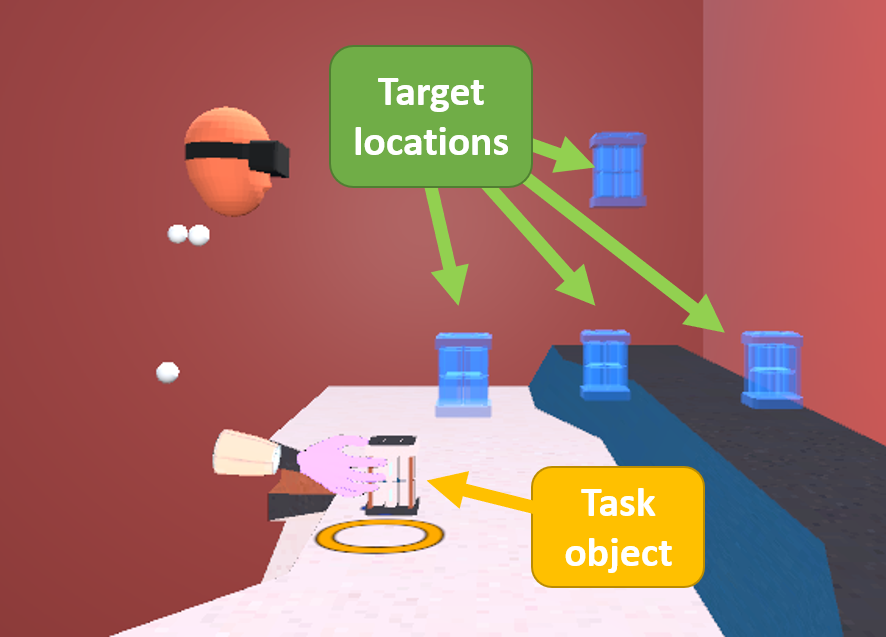}
    \caption{Side view of the task. The object start position and target locations were normalised to the subject's height and arm length.}
    \label{fig:sideView}
\end{subfigure}
~
\begin{subfigure}[t]{0.305\textwidth}
    \includegraphics[width=\textwidth]{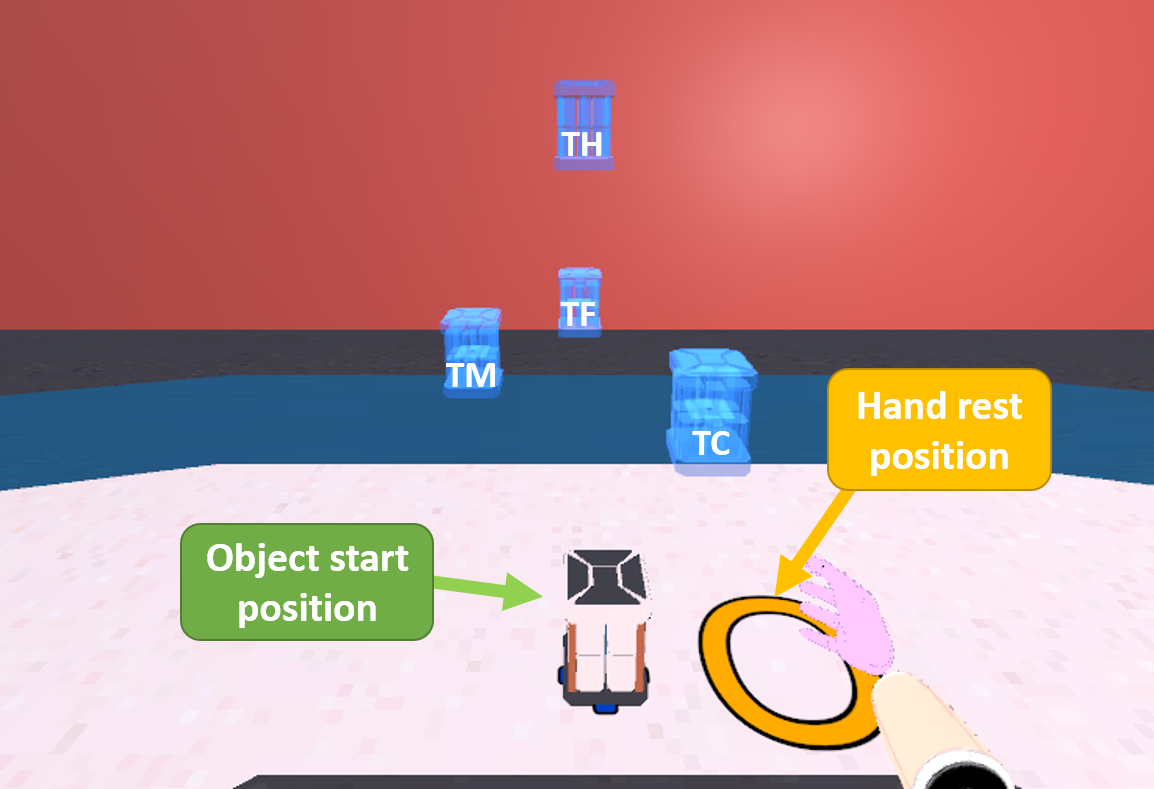}
    \caption{Subject point of view of the task. Only one target location was presented per repetition of the task. Subjects were required to place the virtual hand on the orange circle before the start of the task.}
    \label{fig:subjectView}
\end{subfigure}
~
\begin{subfigure}[t]{0.25\textwidth}
    \includegraphics[width=\textwidth,trim={0 4.5cm 3cm 0},clip]{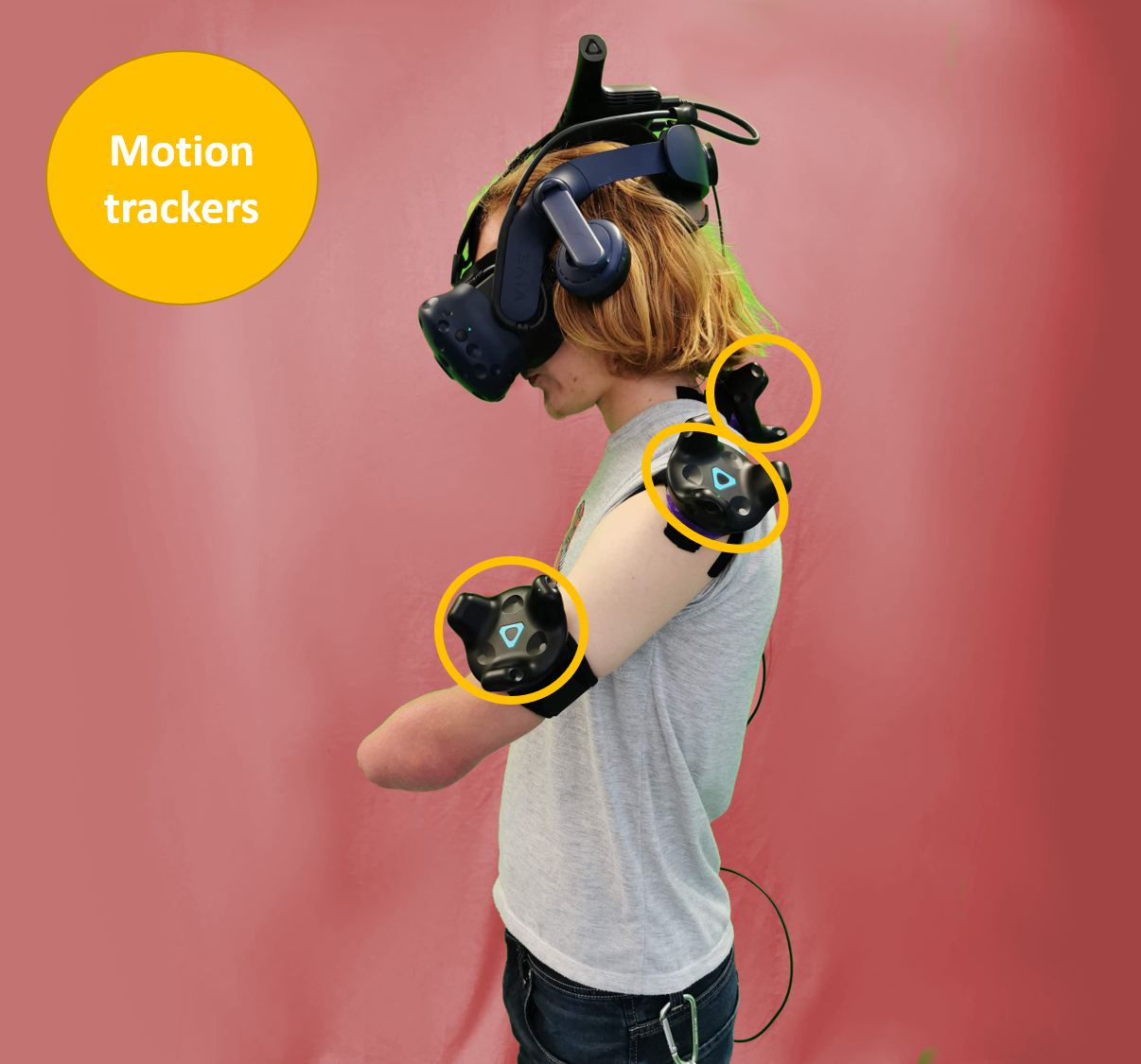}
    \caption{Amputee subject with motion tracking sensors. Sensors were placed on C7 vertebrae, shoulder acromion, and upper arm.}
    \label{fig:amputeeSubject}
\end{subfigure}

\begin{subfigure}[t]{0.30\textwidth}
    \includegraphics[width=\textwidth]{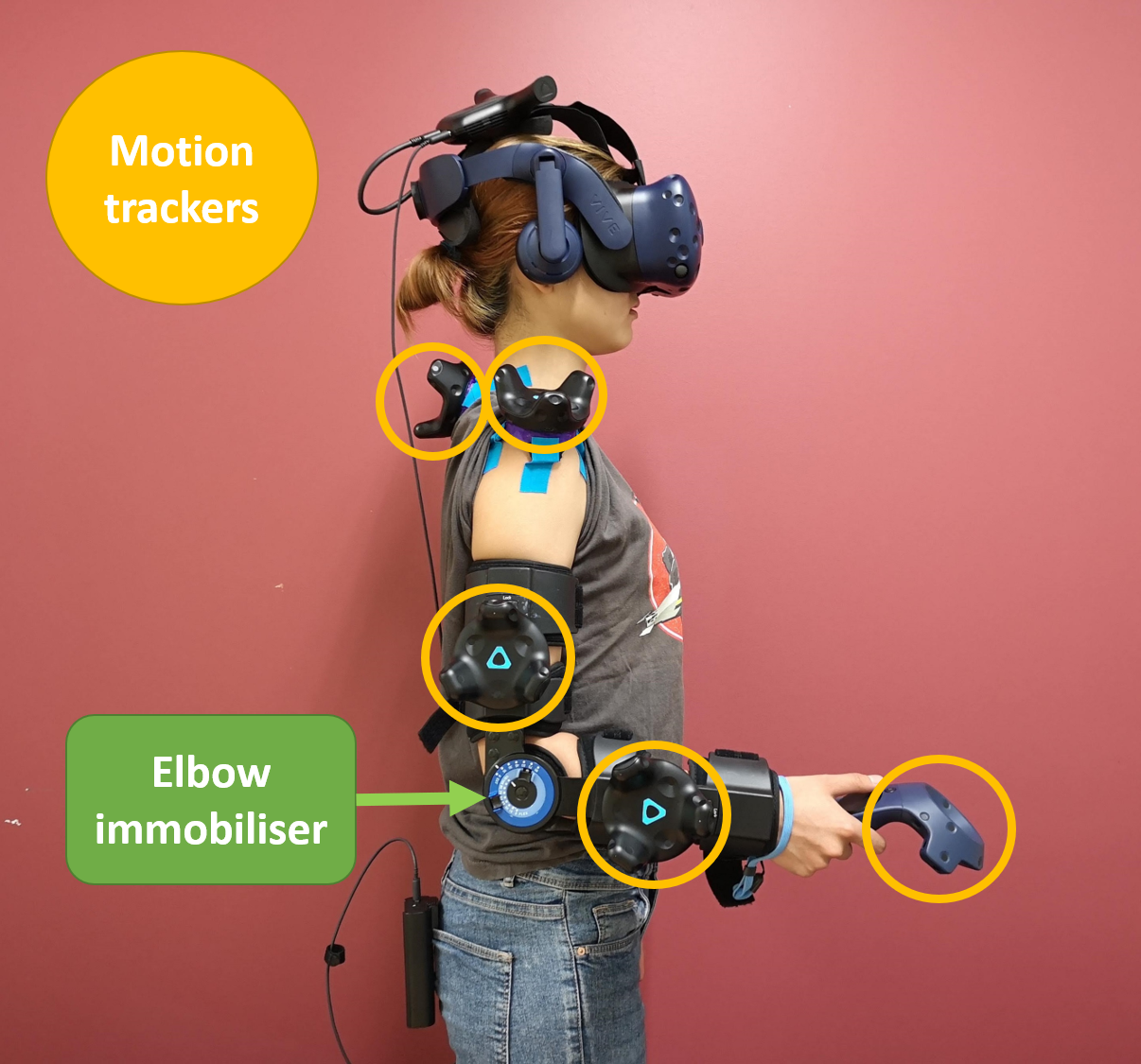}
    \caption{Subject with motion tracking sensors, elbow brace, and VR headset. Sensors were placed on C7 vertebrae, shoulder acromion, upper arm, forearm, and hand. An elbow immobiliser was used for the virtual prosthetic cases. In the EMG case, two sEMG sensors were placed on the subject's forearm.}
    \label{fig:subject}
\end{subfigure} 
~ 
\begin{subfigure}[t]{0.27\textwidth}
    \includegraphics[width=\textwidth]{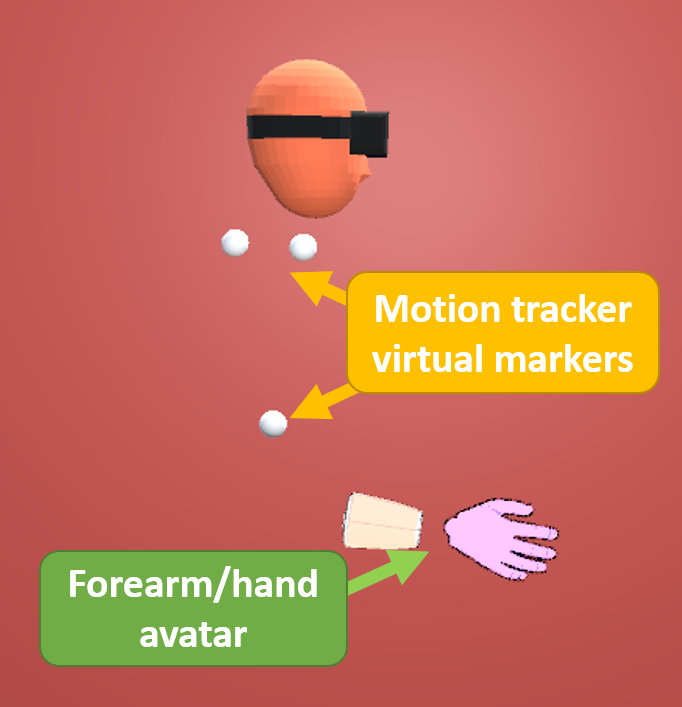}
    \caption{Able-bodied VR avatar. Motion trackers determine the motion of the forearm and hand avatars, and the white markers. The elbow immobiliser was inactive to allow the subject to perform the task naturally.}
    \label{fig:ableAvatar}
\end{subfigure}
~ 
\begin{subfigure}[t]{0.27\textwidth}
    \includegraphics[width=\textwidth]{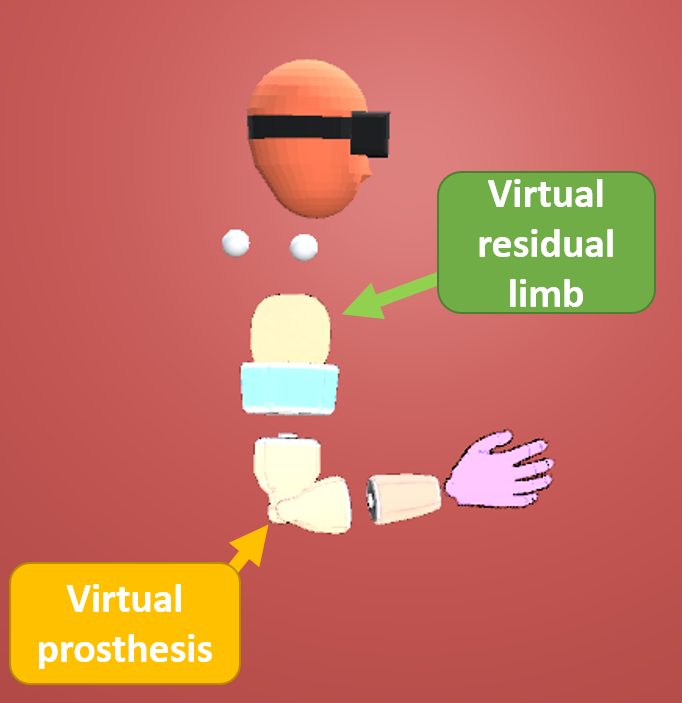}
    \caption{Prosthetic VR avatar. The upper arm motion tracker was used to create a ``residual limb'' avatar to which the virtual prosthesis was attached. The rest of the trackers were presented as white markers. The elbow immobiliser was engaged, and forearm and hand tracking was disabled.}
    \label{fig:prosthesisAvatar}
\end{subfigure}
\caption{Experimental platform and task set-up. (a) presents the side view of the task and (b) the subject point of view in VR. (c) and (d) show the motion tracking sensor placement, and (e) and (f) the VR avatars for the able-bodied and prosthetic cases, respectively.}
\label{fig:taskSetup}
\end{figure*}

An experiment was designed to analyse the performance of the proposed task-space synergy method when used in a basic prosthetic elbow. The resultant behaviour with the proposed task-space synergy-based prosthetic interface was compared with able-bodied behaviour, a traditional sEMG-based prosthetic interface, and a basic joint-space synergy-based prosthetic interface. The experiment was performed in a Head Mounted Display-based (HMD) Virtual Reality Environment (VRE) on able-bodied subjects (Figure \ref{fig:subject}). The use of a VRE allowed the gathering of data from able-bodied subjects for both able-bodied and prosthetic cases by virtually removing their lower arm and fitting them with a virtual prosthesis. The case study with an amputee subject used the same VRE-based methodology. A video description of the VR platform used for the experiment can be accessed from https://youtu.be/vvq4tjShdB4.

%
%
%
%

\subsection{Experiment Task and Set-up}
The task required subjects to pick and place an object in 3D space in the forward direction from a standing position without stepping. The execution of the task is considered as one iteration from the moment the object was grasped until the object was placed. To ensure repeatability of the task, canonical examples containing fixed start and target position were posed. The choice of start (elbow at $90^{\circ}$) and target positions (four different targets) was done to focus the experiment on the reaching phase of motion, which is critical in the use of above-elbow prostheses. One object start position and four target positions were used and were placed at a location relative to each subject's height ($h$) and arm length ($\ell$). The target location details are presented in Table \ref{ta:targets}, these locations are relative to the subject's initial standing position. Figure \ref{fig:sideView} shows the start and target locations in the VRE from a side view, while Figure \ref{fig:subjectView} shows them from a subject point of view.
\begin{table}[ht]
    \caption{Object start and target positions.}
    \centering
    \begin{tabular}{|c||c|c|c|p{3.8cm}|}
        \hline
        Target & $x (m)$ & $y (m)$ & $z (m)$ & Comment \\
        \hline
        \hline
        Start  &     $0.5\ell$     &     $0.5h$   &  0 & Reachable from a neutral, relaxed position with the upper-arm down and elbow bent $90\deg$.\\
        \hline
        Close  &     $0.75\ell$    &     $0.65h$   &  0.12 & Reachable with upper-body motion only.  \\
        \hline
        Mid  &     $1.0\ell$      &     $0.65h$   &  -0.12 & Reachable with either upper-body motion or arm extension only. \\
        \hline
        Far  &     $1.5\ell$      &     $0.65h$   &  0 & Reachable with both upper-body motion and arm extension only.  \\
        \hline
        High  &     $1.0\ell$      &     $0.9h$   &  0  & Reachable with both upper-body motion and arm extension only. \\
        \hline
    \end{tabular}
    \label{ta:targets}
\end{table}

Four different experiment cases were used, one able-bodied and three transhumeral prosthetic. 
\begin{enumerate}
    \item \textbf{AB} (Able-bodied): This case was used to determine the benchmark motor behaviour and task performance for each subject.
    \item \textbf{EP} (sEMG Proportional): A standard dual-site differential surface EMG proportional prosthetic interface was used to command the prosthetic elbow's joint velocity \cite{Fougner2012}. The rectified and thresholded sEMG signals from two antagonistic muscles were used to determine the direction and angular velocity of the prosthetic elbow joint. This case was used as the traditional prosthetic interface for comparison.
    \item \textbf{TS} (Task-space Synergy): The proposed task-space synergy method as described in equation (\ref{eq:taskSpaceSynergy}).
    \item \textbf{JS} (Joint-space Synergy): A Joint-space synergy method (JS) given by a shoulder-elbow flexion synergy was used as the synergistic prosthetic interface case for comparison.
\end{enumerate}

The range of motion of the elbow was limited to $5^{\circ} \leq q_e \leq 140^{\circ}$ to avoid singularities. The JS was given by $\dot{q}_e = \theta \dot{q}_s$; where $\dot{q}_e$ is the prosthetic elbow velocity, $\dot{q}_s$ the shoulder extension velocity, and $\theta = 1$ the synergy parameter. This joint-space synergy, based on \cite{Alshammary2018}, was chosen because it is the closest to the proposed task-space synergy as it does not require training or calibration by allowing the user to compensate for the synergy by choosing when in the reach to enable it.

Subjects were fitted with an elbow orthosis with two motion tracking sensors attached to it, one for the lower-arm and one for the upper-arm.  Two additional motion tracking sensors were placed on the C7 vertebrae (trunk motion), the shoulder acromion (shoulder motion), and the hand (see Figure \ref{fig:subject}). The elbow orthosis allowed to immobilise the subject's elbow in the prosthetic cases to better represent muscle conditions in amputees, as done in \cite{Alshammary2018}. The amputee subject was fitted as able-bodied on his right arm for the AB case, while for the prosthetic cases he had the upper-arm sensor strapped to his left upper arm (see Figure \ref{fig:amputeeSubject}).

In the AB case, motion of the upper-body, the whole arm and hand was directly mapped to the VRE through a virtual avatar as shown in Figure \ref{fig:ableAvatar}. For the case study with the subject with limb difference, his able-bodied (right) arm was used for the AB case. In the prosthetic cases, transhumeral amputation was emulated in able-bodied subjects in the VRE by using the data from the upper-arm motion tracker to determine the motion of the virtual residual limb. A virtual prosthetic elbow, forearm, and hand were attached to the virtual residual limb as seen in Figure \ref{fig:prosthesisAvatar}. For the case study with the subject with limb difference, his arm with limb difference (left) was used for the prosthetic cases. Only the virtual prosthetic elbow was actuated, while the forearm and hand were fixed.

Muscle activation was gathered using sEMG electrodes placed on the forearm as the elbow orthosis interfered with the biceps/triceps electrode placement typically used in transhumeral amputees. While the forearm electrode position does not reflect the physiological placement of sEMG sensors for transhumeral prostheses, it can provide comparable functional capabilities in able-bodied subjects as observed in \cite{Alshammary2018}.

The following prosthetic interface simplifications were done to reduce the cognitive load on the subjects and to focus the experiment on the motor behaviour aspect of prosthesis use. Firstly, the functionality of the prosthetic elbow was enabled in a toggle fashion by a press of the centre button (thumb) of the Vive controller (Figure \ref{fig:subject}), which was held in the able-bodied hand of the amputee subject. This was done to avoid relying on complex sEMG switching commands, which are a major issue with active prostheses to date \cite{Biddiss2007a}. Secondly, grasping and releasing of the object were done automatically when the hand stopped moving and was touching the object (grasp) or within the target area (release), avoiding the need to switch between elbow and hand control.

\subsection{Hardware Set-up}
The experiment was performed on an HTC Vive Pro HMD with the application developed in Unity3D. The experimental platform runs on an Intel Core i7-8700K processor at 3.7GHz, with 32GB RAM, and an Gigabyte GeForce GTX 1080Ti video card with 11GB GDDR5. HTC Vive Trackers were used for motion tracking, except for hand motion tracking where an HTC Vive Controller was used (HTC Vive system tracking capabilities can be found in \cite{Niehorster2017}). The C7 and shoulder acromion Vive Trackers were attached directly to the subject's body using medical tape. Data gathering, and VR update were performed at 90Hz. Myoware sensors were used for sEMG data gathering with Ag-AgCl electrodes. The VR platform used for the experiment can be downloaded from https://github.com/Rigaro/VRProEP.

\subsection{Experimental Protocol}
The experiment was performed on 15 able-bodied subjects, six female and nine male. The age range was $[22, 34]$ and the median age $28$. The amputee case study was performed on a single male subject aged $23$. The subject has a congenital limb difference in his left forearm as seen in Figure \ref{fig:amputeeSubject}. He is not a prosthesis user and does not use his left arm in daily tasks relevant to the task tested in this experiment. The experiment was performed in two sessions lasting two hours each, on two different days not more than a week apart. Each session tested two cases. The first case tested for all subjects was AB which was used as the benchmark for individual motor behaviour. The order of the following cases was randomly selected from the three cases representing prosthetic-use (EP, TS or JS) to minimise the bias introduced by the effects of learning on the data gathered. The session procedure is presented in Table \ref{ta:protocol}.
\begin{table}[ht]
    \caption{Experiment session protocol.}
    \centering
    \begin{tabular}{|c||c|c|}
        \hline
        Step & Time (minutes) & VR/No-VR \\
        \hline
        \hline
        Introduction   &     5     &     No-VR \\
        \hline
        Sensor placement   &     15   &     No-VR    \\
        \hline
        Case 1  &     50      &     VR  \\
        \hline
        Break   &     5      &     No-VR  \\
        \hline
        Case 2  &     50      &     VR  \\
        \hline
    \end{tabular}
    \label{ta:protocol}
\end{table}

Each block of case testing consisted of 120 iterations of the task separated into three sets of 40 iterations, the procedure is presented in Table \ref{ta:caseProtocol}. During the Training block of the experiment, subjects were instructed outside the VRE on how to use the prosthetic interface and were allowed to familiarise themselves with it. This was done by presenting a third person view of the virtual residual limb and prosthetic device on a screen. The procedure was approved by the University of Melbourne Human Research Ethics Committee, project number 1750711.2. Informed consent was received from all subjects in the study.
\begin{table}[ht]
    \caption{Case testing block protocol.}
    \centering
    \begin{tabular}{|c||c|c|c|}
        \hline
        Step & Task iterations per target & Time (minutes) & VR/No-VR \\
        \hline
        \hline
        Training   &     -    &     5    &     No-VR \\
        \hline
        Practice   &     5    &     -    &     VR    \\
        \hline
        Set 1      &     10   &     -   &     VR  \\
        \hline
        Break   &     -      &     1    &     VR  \\
        \hline
        Set 2      &     10   &     -   &     VR  \\
        \hline
        Break   &     -      &     1    &     VR  \\
        \hline
        Set 3      &     10   &     -   &     VR  \\
        \hline
    \end{tabular}
    \label{ta:caseProtocol}
\end{table}

\subsection{Data Gathering and Analysis}
The data gathered from motion tracking sensors and the VRE was used to analyse the performance of subjects in terms of the task, quality of motion, and the resulting motor behaviour. Given that target locations were personalised to each subject, hand trajectories were normalised to the subject's arm length for all cases and targets. This was done by applying a spatial normalisation to all the trajectories. The resulting normalised reaching paths had starting position coordinates $(0,0)$, target position coordinates $(1,0)$, and the trajectories occurred on a plane. The normalised trajectories were uniformly resampled to 200 samples to isolate the spatial and temporal components of the trajectories. An FIR anti-aliasing low-pass filter of order five with a Kaiser window of 20 and linear interpolation were used for resampling. The normalisation procedure follows the methods described in \cite{Yang2002}. The metrics used for data analysis are presented next.


Two metrics of task performance were used: completion time and reach terminal error. The median task completion time ($t_f^{\mu}$) over iterations of the task is used to determine the effects of prosthetic interfaces on the time required to complete the task. As this metric is commonly used in the literature, it is used to compare the performance of the synergistic interfaces in this paper to other studies in the literature. The mean of the reach terminal error ($\delta^{\mu}$) over iterations of the task is used to verify that the task is being performed adequately. Reach terminal error is calculated as $\delta^{\mu}=\|\pv^{i}(t_f) - \pv^{i}_{d}\|_2$, where $\pv^{i}(t_f)$ is the hand position at the end of the reach and $\pv^{i}_{d}$ is the desired hand position for the given target.

Two metrics of quality of motion were used: path smoothness and variability. Continuous, well coordinated movements are characteristic of trained and healthy motor behaviour. Hand and joint path smoothness are well documented and accepted measures of the quality of upper-limb motion \cite{Kistemaker2014, Balasubramanian2012b, DeLosReyes-Guzman2014}. Here Spectral Arc Length (SAL), represented by ($\eta$), is used as the metric for smoothness \cite{Balasubramanian2012a} due to its characteristics which include sensitivity to changes in motor characteristics, robustness to measurement noise, and dimensionless monotonic response. SAL measures the arc length of the curve generated by the Fourier magnitude spectrum of the velocity trajectory to quantify movement smoothness \cite{Balasubramanian2012a}. Low path variability, both of the hand and joints, is characteristic of trained and healthy motor behaviour \cite{Balasubramanian2012b, Darling1987}. Here the mean standard deviation of the hand and joint paths across the time domain ($\sigma_{\mu}$), is used as the metric for path variability. The standard deviation of the paths for each subject, modality and target is calculated at sample across the time domain using the normalised paths, and the mean over time of the standard deviation is used as the metric $\sigma_{\mu}$.

Two metrics of motor behaviour were used: hand/joint path difference and upper-body displacement. The first metric of motor behaviour is path difference ($\Delta_{\mu}$). This metric evaluates the difference between the mean able-bodied hand/joint path and the mean prosthetic hand/joint path. It is used to determine the closeness of the resultant prosthetic motor behaviour to the individual's natural able-bodied motor behaviour. Let the mean hand/joint path for the able-bodied case ($AB$) be given by $\xv_{AB}^{\mu}(n t_s)$ and the mean path for the prosthetic case ($PC$) given by $\xv_{PC}^{\mu}(n t_s)$, where $t_s$ is the sampling time and $n\in\Natural$. The resampled normalised paths are used to calculate the mean paths $\xv_{AB}$ and $\xv_{PC}$, such that both $AB$ and $PC$ cases have the same normalised start and end positions, and sample length. The path difference ($\Delta_{\mu}$) is defined as the sum of the Euclidean distances between the mean paths across the time domain as follows:

\begin{equation}
\label{eq:handDiff}
    \Delta_{\mu} =  \sum_{n=0}^{N} \| \xv_{AB}^{\mu}(n t_s)- \xv_{PC}^{\mu}(n t_s) \|_2 .
\end{equation}
    
The second metric of motor behaviour is upper-body displacement ($\phi^{\mu}$). Upper-body displacement is calculated as the Euclidean distance between in initial and final positions of the C7 and shoulder sensors. This metric evaluates the amount of upper-body motion used for the task. When a prosthetic device is not used effectively or is not adequate for the task this metric represents compensation motion. Some level of upper-body motion is also part of natural able-bodied motion during reaching tasks.

Statistical analyses of the of these metrics were performed in Minitab 19. The last 10 data-points from each modality and target (last set of task iterations) were chosen for the analyses in order to minimise the effects of human motor learning on the results. The data for all metrics, except smoothness and shoulder displacement, was transformed to a logarithmic scale to satisfy the assumptions of the general linear model, in particular, the constant variance for the random error. The data was fitted to a general linear model using both the subjects and modalities as model factors. The metrics presented previously were used as the dependent variables. The general linear model was used to perform two-way analyses of variance (two-way ANOVA) and comparisons between the modalities with a confidence level of 95\% ($p<0.05$) after adjustment for multiple comparisons using Tukey's method. The analyses were used to determine whether statistically significant differences exist between the modalities in terms of the evaluation metrics. Statistical analyses were performed in consultation of the University of Melbourne Statistical Consulting Platform.

%% file: 6_results_v3.tex
\section{Results \& Discussion}
Results for the experiment with able-bodied subjects and the case study with an amputee subject are presented next. The data-set of the results (raw and processed) can be downloaded from https://github.com/Rigaro/TaskSpaceSynergyData2020.

\subsection{Experiment with Able-bodied Subjects}

\begin{figure}[htb]
\centering
\includegraphics[trim={0pt 0pt 0pt 0pt}, width=0.85\columnwidth, clip]{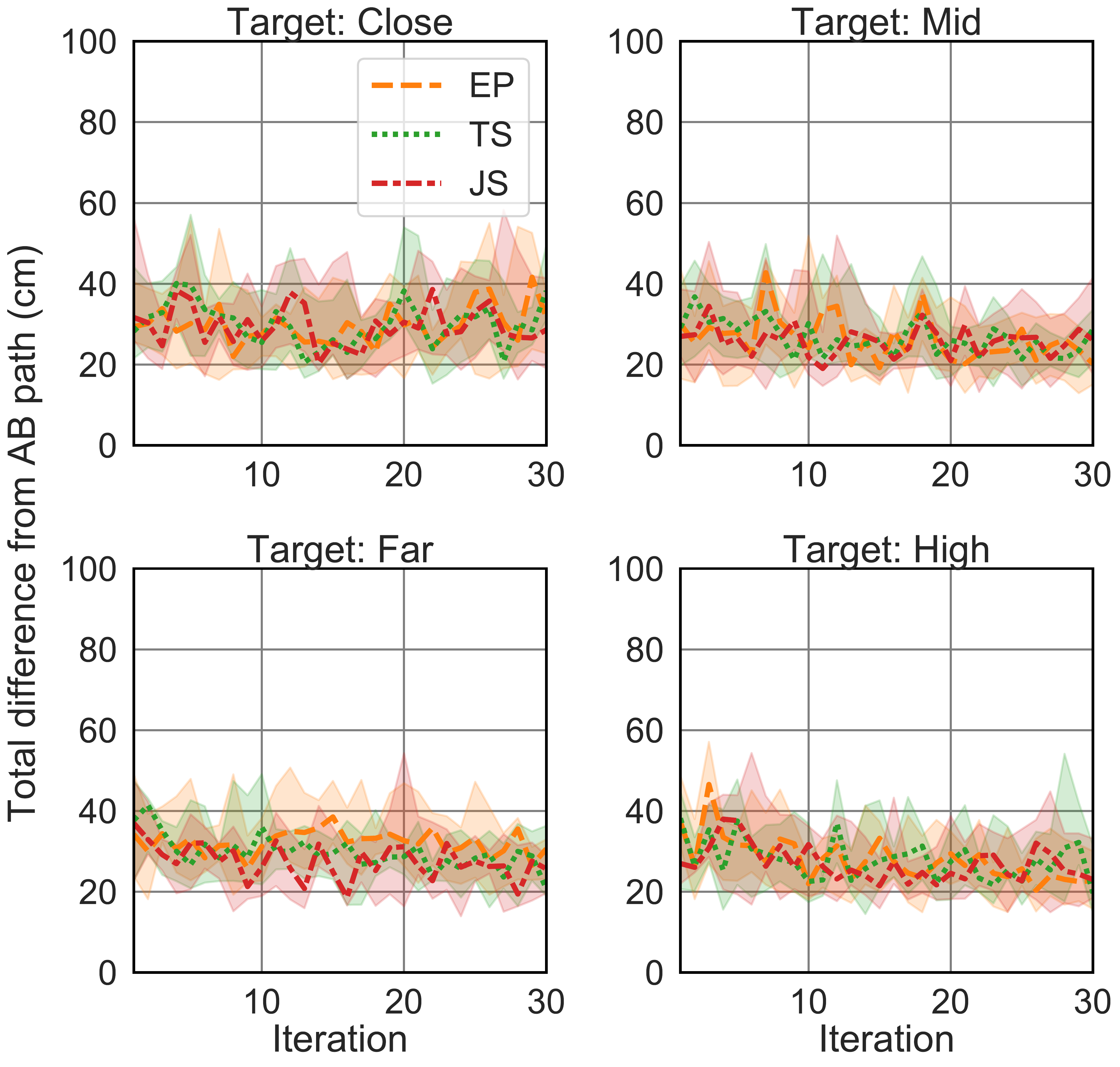}
\caption{Population median experimental results for hand path difference from AB ($\Delta_\mu$) over iterations of the task for each target.}
\label{fig:handPathDiff}
\end{figure}

The results for the hand path difference over iterations of the task are presented in Figure \ref{fig:handPathDiff}, where the population median of the hand path difference for each iteration of the task is presented for each target. 

From these results, it can be seen that as subjects repeat the task the difference between their able-bodied hand path and the prosthesis hand path is reduced. In general, it can be observed that there was a gradual reduction of the difference over the first 10 iterations after which the progress slowed down.

The reduction in the path difference over iterations suggests that individuals try to reproduce able-bodied motion with the prostheses. Thus, leading to an improvement on their performance. The clear difference from the AB path shows that despite their best effort, the introduction of limb loss and the artificial limb inherently changes the dynamics of reaching. This means that the natural/dynamically optimal motion when using a prosthesis is different from one's AB motion. The comparisons test showed no statistically significant performance differences across the three modalities. These results suggest that even if the hand paths followed by each prosthetic modality have their differences, the total difference to the able-bodied hand path is similar.

\begin{figure}[ht]
\centering
\includegraphics[trim={0pt 0pt 0pt 0pt}, width=0.85\columnwidth, clip]{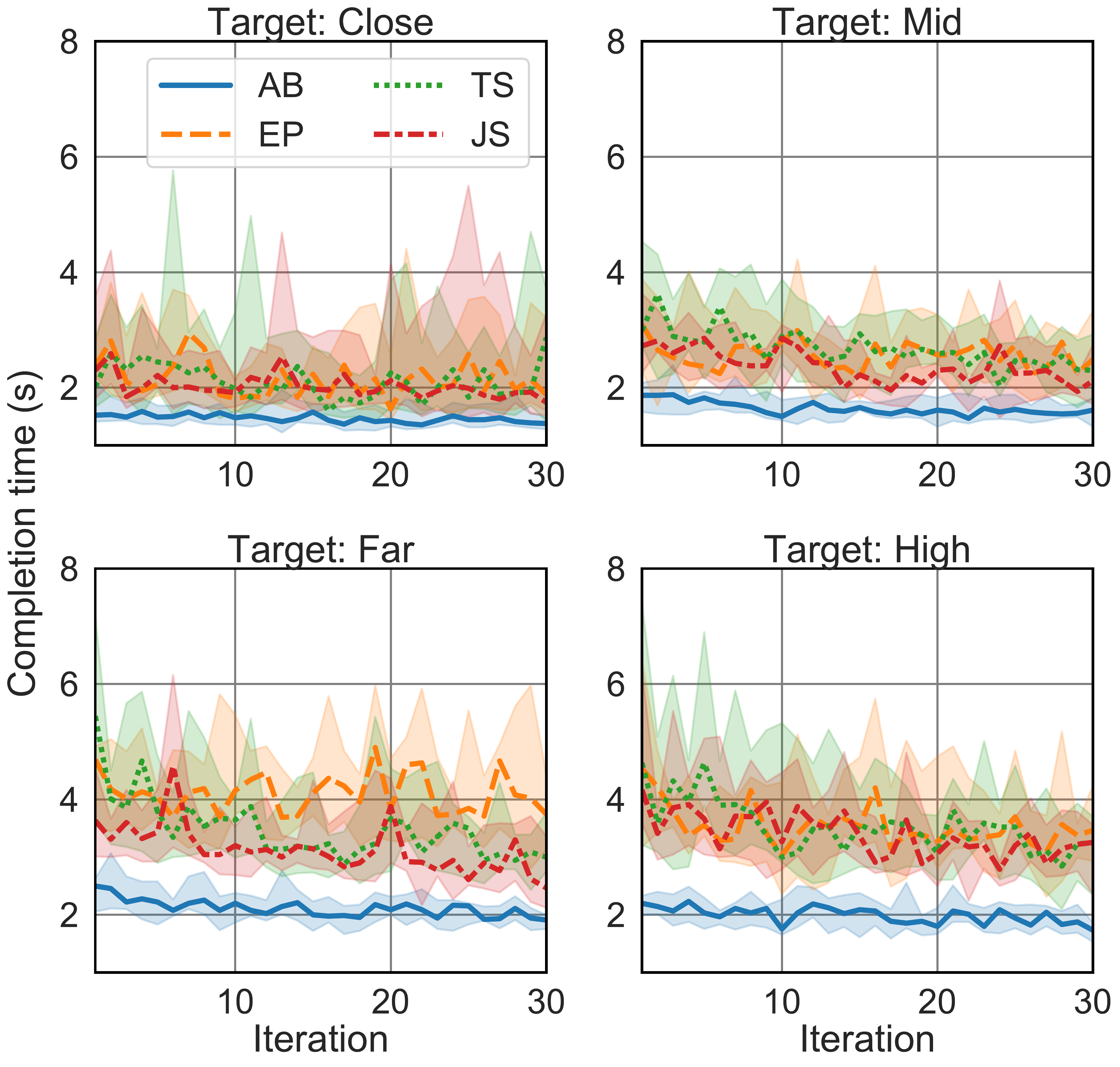}
\caption{Population median experimental results for task completion time ($t_f^{\mu}$) over iterations of the task for each target.}
\label{fig:task time}
\end{figure}

The improvement in task performance can be observed in the task completion time, shown in Figure \ref{fig:task time}, which is reduced over iterations of the task. 

The statistical analysis showed that there were significant differences between the synergistic interfaces (TS and JS) and EP. Synergistic interfaces had on average a 20\% shorter task completion time than sEMG for the Far target, which requires more joints to be coordinated (upper-body and upper-arm). Moreover, no statistically significant difference between TS and JS was found. This suggests that TS may be as effective as JS in reducing the time required to perform reaching tasks. While in \cite{Alshammary2018} it was reported that JS had a significant advantage over EP in terms of completion time, from the results there it was not possible to conclude whether it was because of the motion-based prosthetic joint-switching mechanism proposed therein or the synergy (JS) itself. The results presented herein suggest that the significant difference between EP and JS reported in \cite{Alshammary2018} may be due to the switching mechanism, as a major difference is only observed for the reaching task that requires more joints to be coordinated.

There was no statistically significant difference between the results for the reach terminal error metric ($\delta^{\mu}$) across all targets and modalities, with the subject's hand end position being within $4cm$ of the desired target position. This verifies that subjects were capable of performing the task adequately and with similar final position error through the experiment.

\begin{figure*}[htb]
\centering
    \begin{subfigure}[t]{0.4\textwidth}
        \centering
        \includegraphics[trim={20pt 0pt 60pt 0pt}, width=\columnwidth, clip]{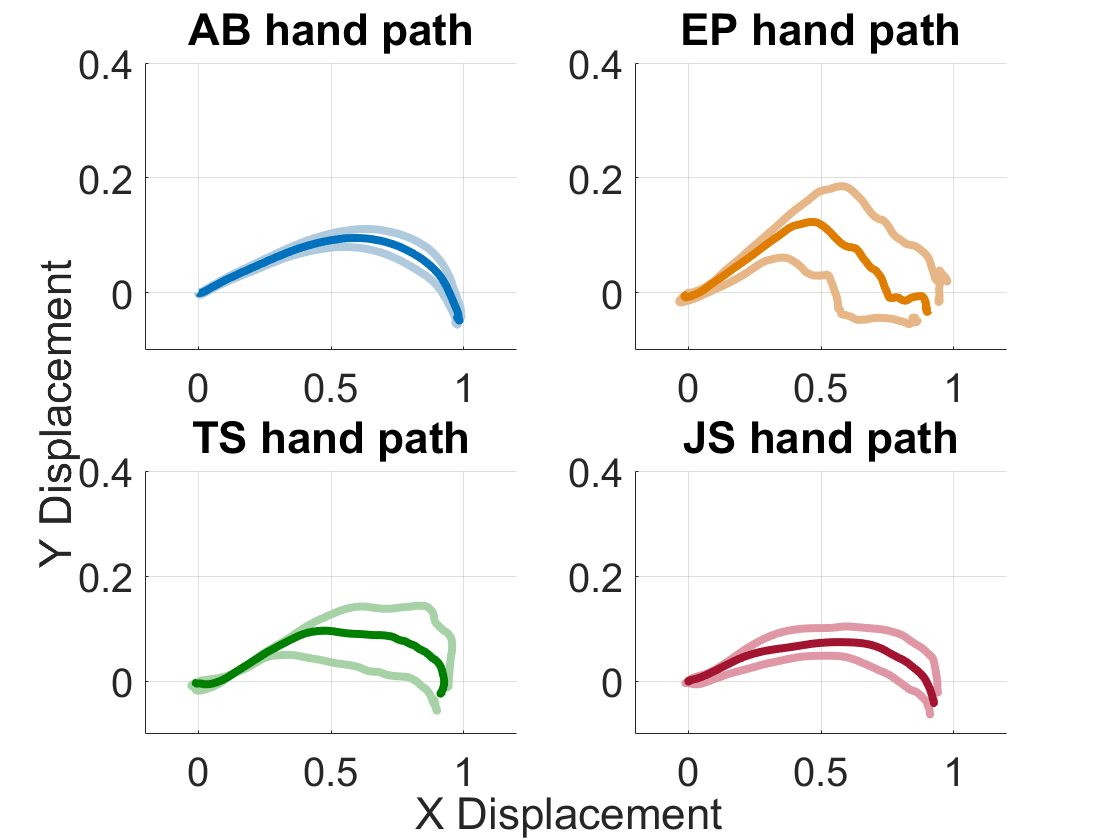}
        \caption{Mean and standard deviation hand paths.}
        \label{fig:repHandPath}
    \end{subfigure}
    ~
    \begin{subfigure}[t]{0.4\textwidth}
        \centering
        \includegraphics[trim={20pt 0pt 50pt 0pt}, width=\columnwidth, clip]{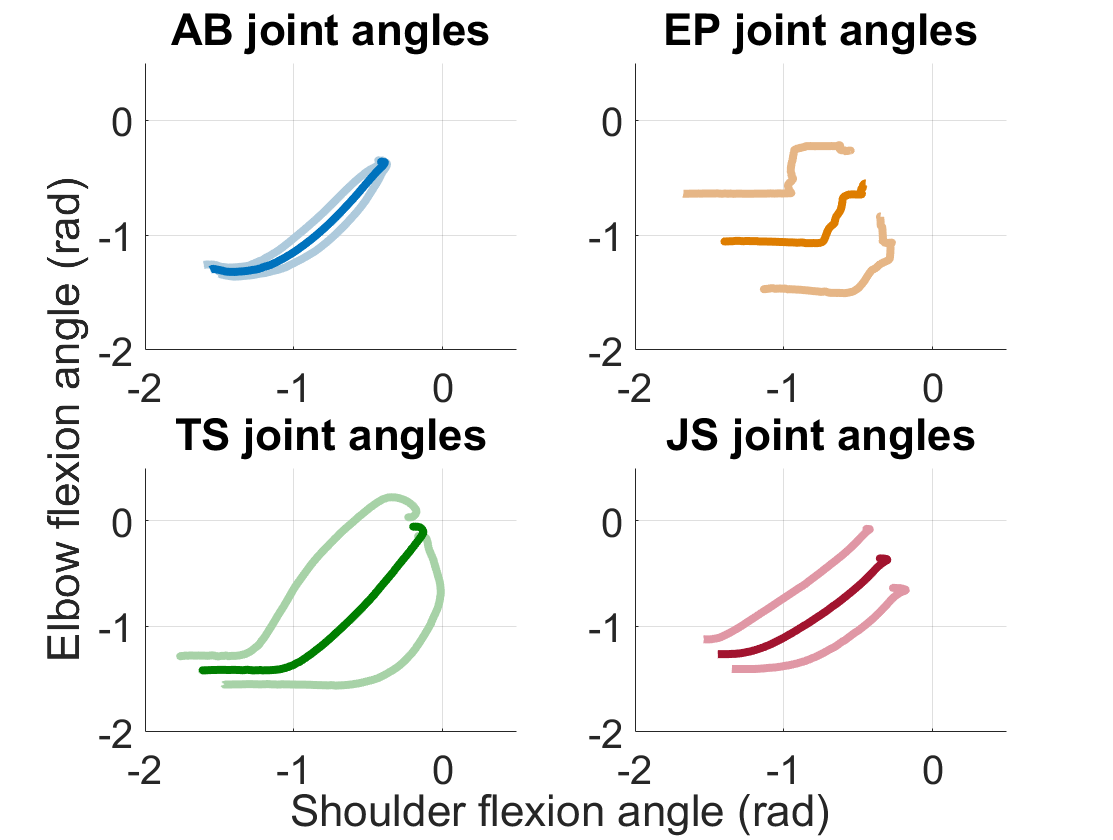}
        \caption{Mean and standard deviation joint paths.}
        \label{fig:repJointSyn}
    \end{subfigure}
    \vspace{-2pt}
    \caption{Mean and standard deviation hand paths and joint angles of the last 10 iterations (steady-state) of the Far target for a representative subject}
    \label{fig:repResults}
\end{figure*}

\begin{figure*}[htb]
\centering
    %
    \begin{subfigure}[t]{0.29\textwidth}
        \centering
        \includegraphics[width=1.0\textwidth]{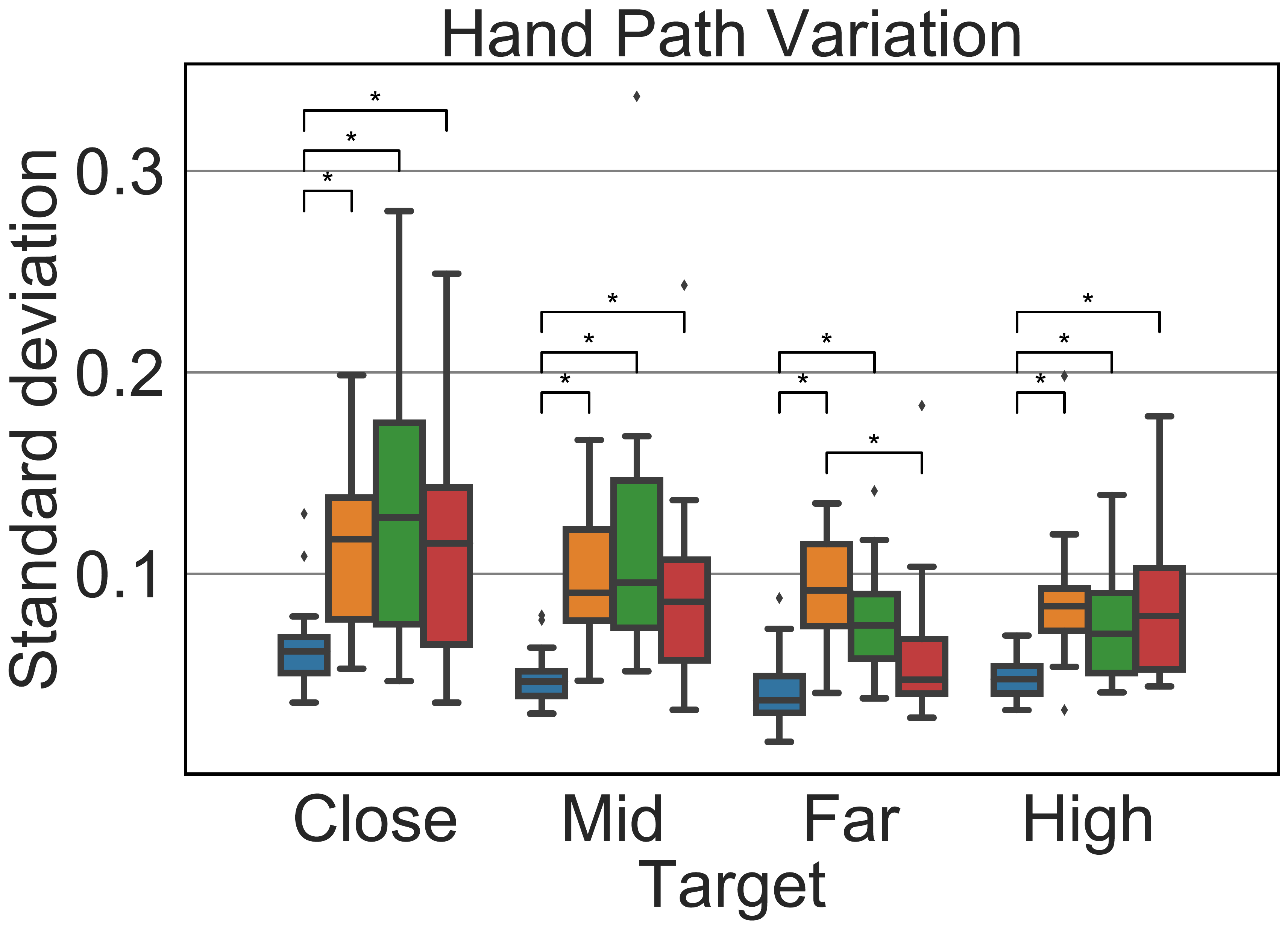}
        \caption{Population hand path variability ($\sigma_{\mu}$).}
        \label{fig:handVar}
    \end{subfigure}
    ~
    %
    \begin{subfigure}[t]{0.29\textwidth}
        \centering
        \includegraphics[width=1.0\textwidth]{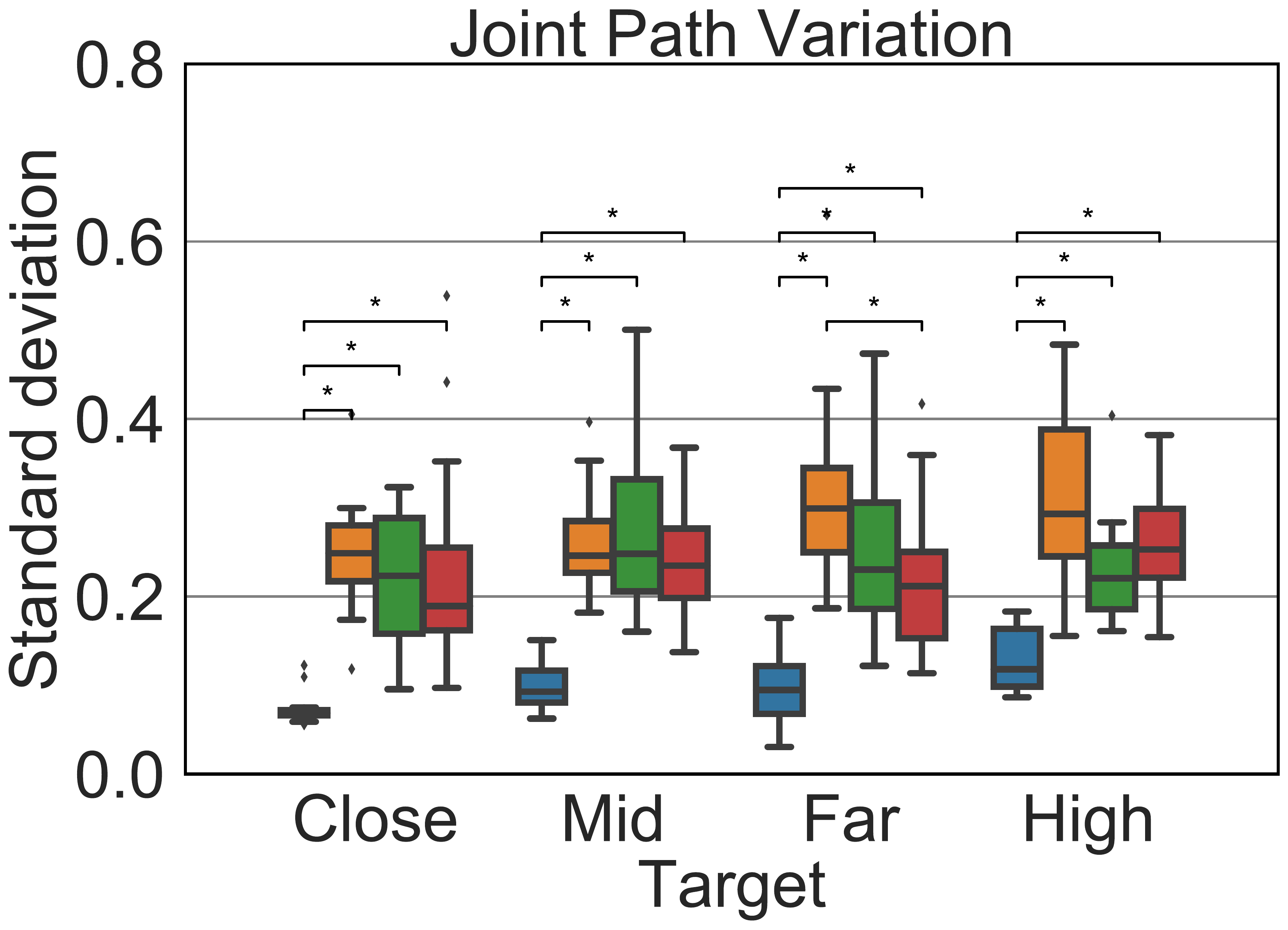}
        \caption{Population joint path variability ($\sigma_{\mu}$).}
        \label{fig:jointVar}
    \end{subfigure}
    ~
    %
    \begin{subfigure}[t]{0.29\textwidth}
        \centering
        \includegraphics[width=1.0\textwidth]{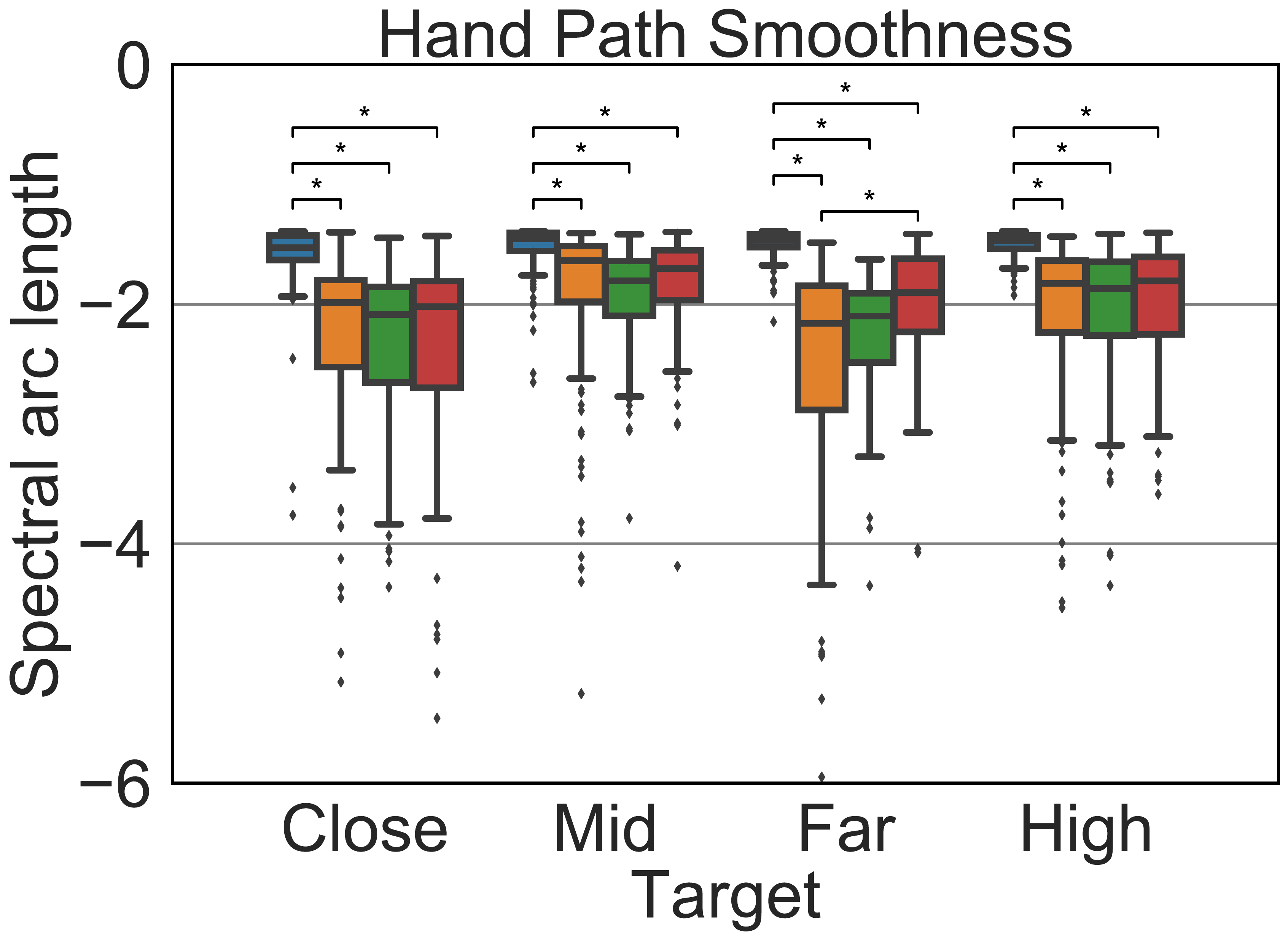}
        \caption{Population hand path smoothness ($\eta$). The less negative the smoother the path.}
        \label{fig:handSAL}
    \end{subfigure}
    
    %
    \begin{subfigure}[t]{0.29\textwidth}
        \centering
        \includegraphics[width=1.0\textwidth]{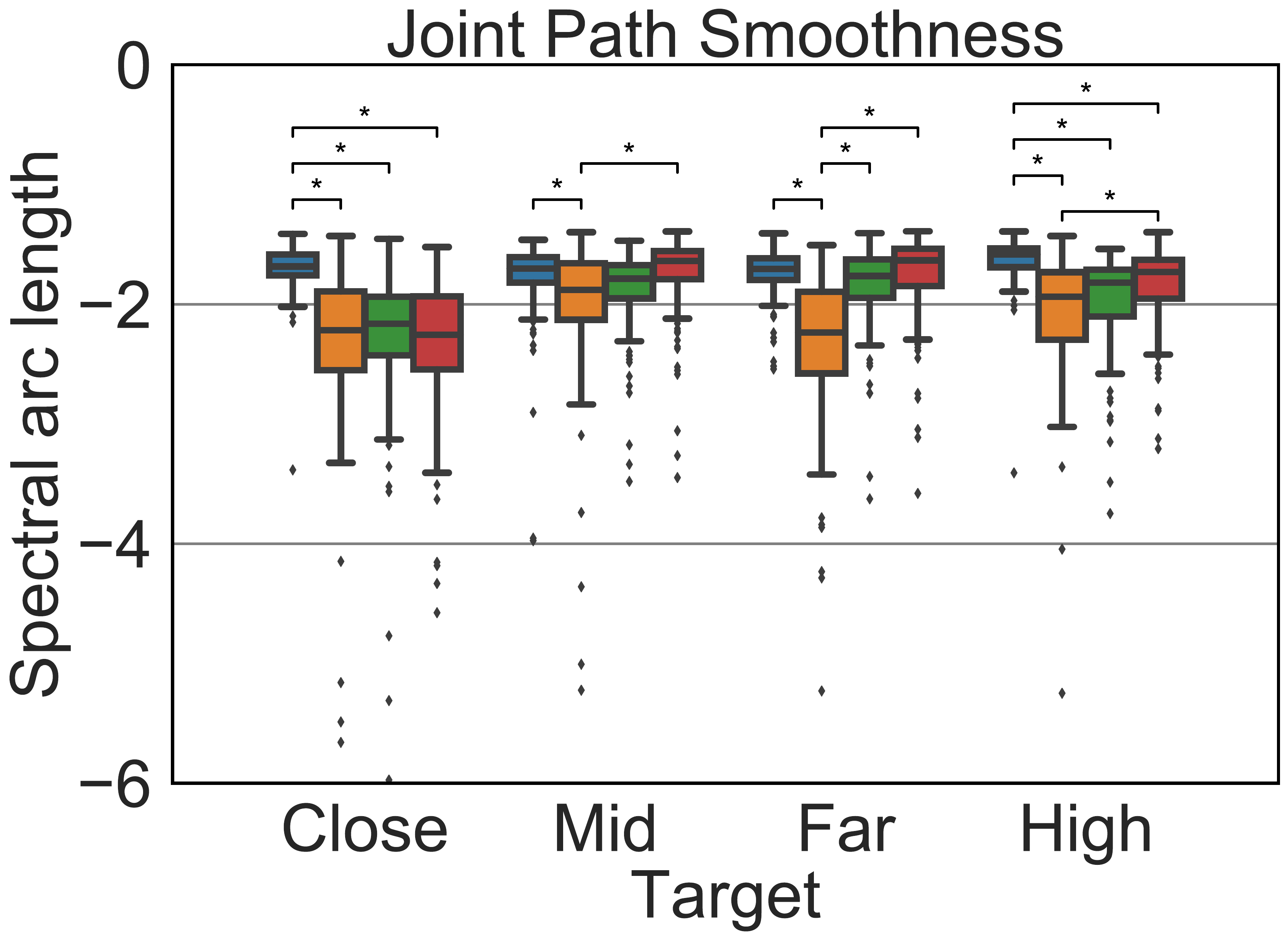}
        \caption{Population joint path smoothness ($\eta$). The less negative the smoother the path.}
        \label{fig:jointSAL}
    \end{subfigure}
    ~
    %
    \begin{subfigure}[t]{0.29\textwidth}
        \includegraphics[width=1.0\textwidth]{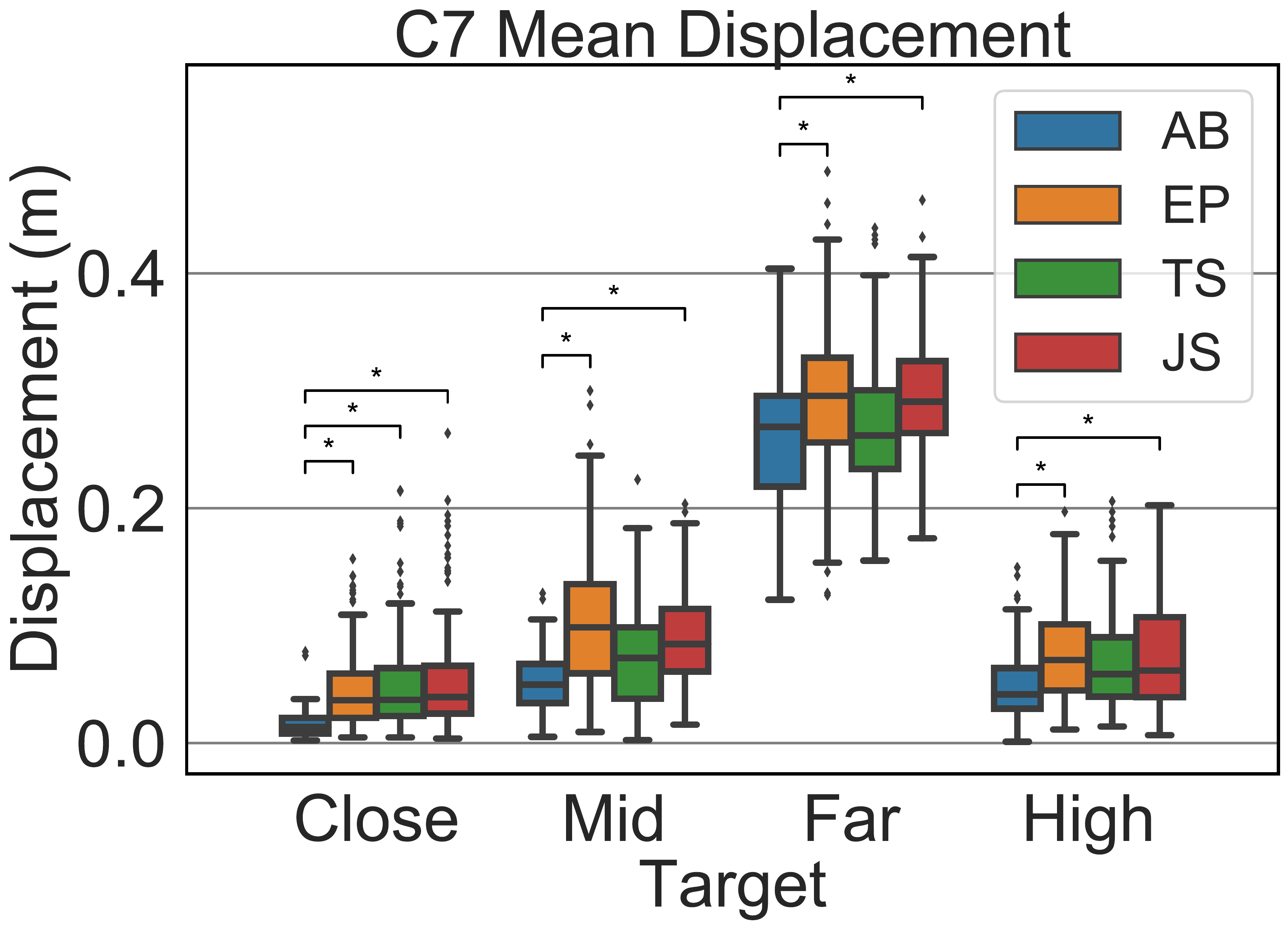}
        \caption{Population C7 displacement ($\phi^{\mu}$).}
        \label{fig:C7Displacement}
    \end{subfigure}
    ~
    \begin{subfigure}[t]{0.29\textwidth}
        \includegraphics[width=1.0\textwidth]{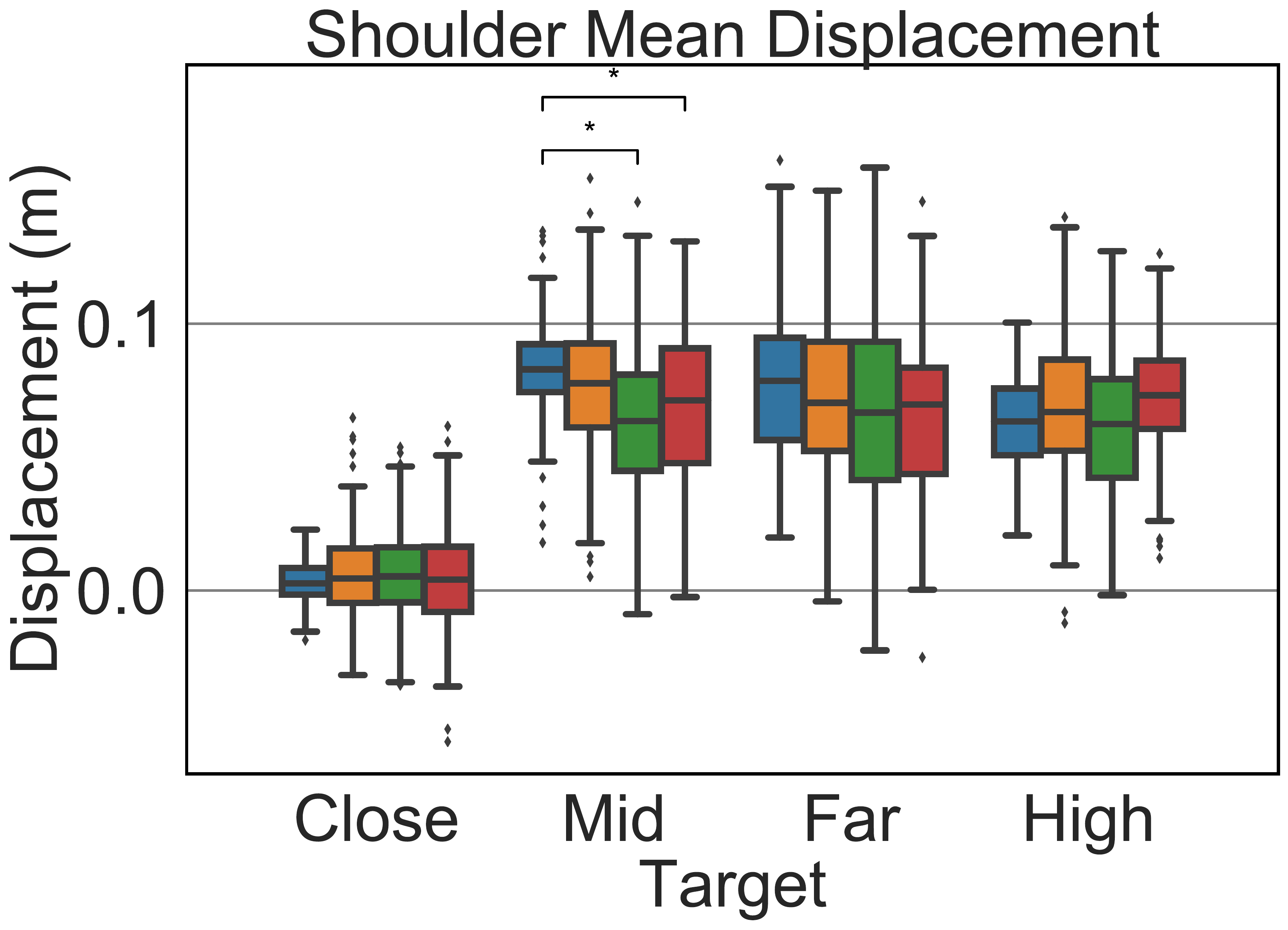}
        \caption{Population shoulder acromion displacement ($\phi^{\mu}$).}
        \label{fig:SADisplacement}
    \end{subfigure}
\caption{Population experimental results for the Quality of Motion and Motor Behaviour metrics for the last 10 iterations per target. Statistically significant differences between modalities ($p < 0.05$) are shown by asterisks (*).}
\label{fig:results1}
\end{figure*}

Given that the results for the Far target show the most significant difference between modalities, the hand and joint path analysis will focus on that target. The mean and standard deviation hand paths and joint angles of the last 10 iterations (steady-state) of the Far target for a representative subject are presented in Figure \ref{fig:repResults}. Figure \ref{fig:results1} presents the population mean hand path and joint path variation, hand path and joint-space smoothness, and trunk and shoulder displacement results.

\subsubsection{Reaching Motion Variability and Smoothness}
The hand path in Figure \ref{fig:repHandPath} was normalised between zero and one, representing the start of the movement and the target location, respectively. From the TS results, the initial aiming movement can be identified from the low variability part of the reach, as this is only performed with the residual limb (shoulder) and thus is less variable. This is followed by the straight hand forward movement generated by the synergy, which has higher variability. The aiming phase can also be seen in the joint movement, as shown in Figure \ref{fig:repJointSyn}, where the mean and standard deviation of shoulder and elbow angles are presented. The TS movement starts with a shoulder-only aiming motion, followed by a smooth curve. This is further corroborated by the variability and smoothness results, presented in Figure \ref{fig:results1}. 

JS showed statistically significant lower hand path variability over EP for the Far target (31\%) while showing no statistically significant difference to AB. While TS showed a 16\% reduction in variability over EP, this difference was not statistically significant. Similarly, JS showed a statistically significant reduction of 31\% in joint path variability over EP for the Far target. TS saw and average reduction of 20\% over EP, though it was not statistically significant. However, no statistically significant difference was found between TS and JS. All other targets showed no statistically significant difference between the prosthetic modalities for both variability metrics.

In terms of hand path smoothness the JS interface shows a statistically significant improvement of 10\% for the Far target with respect to EP. No statistically significant difference between JS, and EP and TS was found. This may be attributed to the need with TS to pre-position the hand before executing the movement, which affects the resultant smoothness metric. On the other hand, joint motion smoothness for the JS case was found statistically significantly improved over EP by 18\%, 24\%, and 15\% for the Mid, Far, and High targets, respectively; while no statistically significant difference to AB was found. On the other hand, the TS case only shows a statistically significant improvement in joint smoothness of 17\% for the Far target with respect to EP. No statistically significant difference between TS and JS was found for the Mid, Far, and High targets.

The results for hand path difference, variation, and smoothness suggest that even though the difference between able-bodied and prosthetic hand paths is similar across all modalities, these paths have different qualities. These results suggest that synergistic modalities improve the prosthetic hand trajectories; however, further exploration of the differences between able-bodied and prosthetic hand trajectories may provide further insights into the advantages synergistic modalities, and what aspects require further improvement.

The joint path of subjects with the EP interface, in Figure \ref{fig:repJointSyn}, shows a mostly horizontal line representing the initial shoulder only movement, followed by a combined elbow and shoulder movement where subjects use EMG to extend the elbow while using the shoulder to do the fine adjustments of hand position. Lastly, JS shows a mostly smooth curve as it was identified that subjects learned the timing to activate the synergistic movement to minimise the error at the end of the motion. Interestingly, this resulted in more natural and joint paths, further demonstrating the capabilities of human motor learning and adaptation. Overall, it appears that regardless of the type of synergy humans will adapt to the best of their capabilities to use the synergy effectively.

\subsubsection{Joint Range and Trunk Movement}
Another important feature that arises from Figure \ref{fig:repJointSyn} is the range of joint angles used. It can be observed that out of all cases, TS utilises a wider range of arm joint angles suggesting more arm utilisation. This is corroborated with the trunk and shoulder displacement metrics, shown in Figures \ref{fig:C7Displacement} and \ref{fig:SADisplacement}, which show that TS has the lowest trunk and shoulder forward displacements. While neither TS nor JS were found to be statistically significantly different to EP in terms of compensation motion, no statistically significant difference between TS and AB was found for the Mid, Far, and High targets. Nevertheless, trunk displacement for the TS case shows a 60\% and 45\% reduction over EP and JS, respectively, for the Mid target; while a 60\% reduction over EP and JS for the Far target. These results agree with those reported in \cite{Merad2018}, which present a fully calibrated synergy, where a reduction in trunk compensation motion was reported with a JS method over EP, suggesting that TS may further reduce compensatory movement at the expense of a less natural movement (point and reach). However, this is due to the constraint imposed on the reaching path. If the location of the reaching target is known in the formulation of the task-space synergy, then it may be possible to generate a human-prosthesis motion that closely resembles able-bodied behaviour, while retaining the user's control over the movement. More importantly, as the task-space synergy is based on the arm kinematics, no time consuming individual synergy training or personalisation is required, and achieves comparable performance to a joint-space synergy. Moreover, task-space synergies could be generalised to any point-to-point reaching task. Therefore, this scenario will be explored in future studies.

%
%
\subsection{Case Study with Amputee Subject}
\begin{figure}[htb]
\centering
\includegraphics[trim={0pt 0pt 0pt 0pt}, width=0.85\columnwidth, clip]{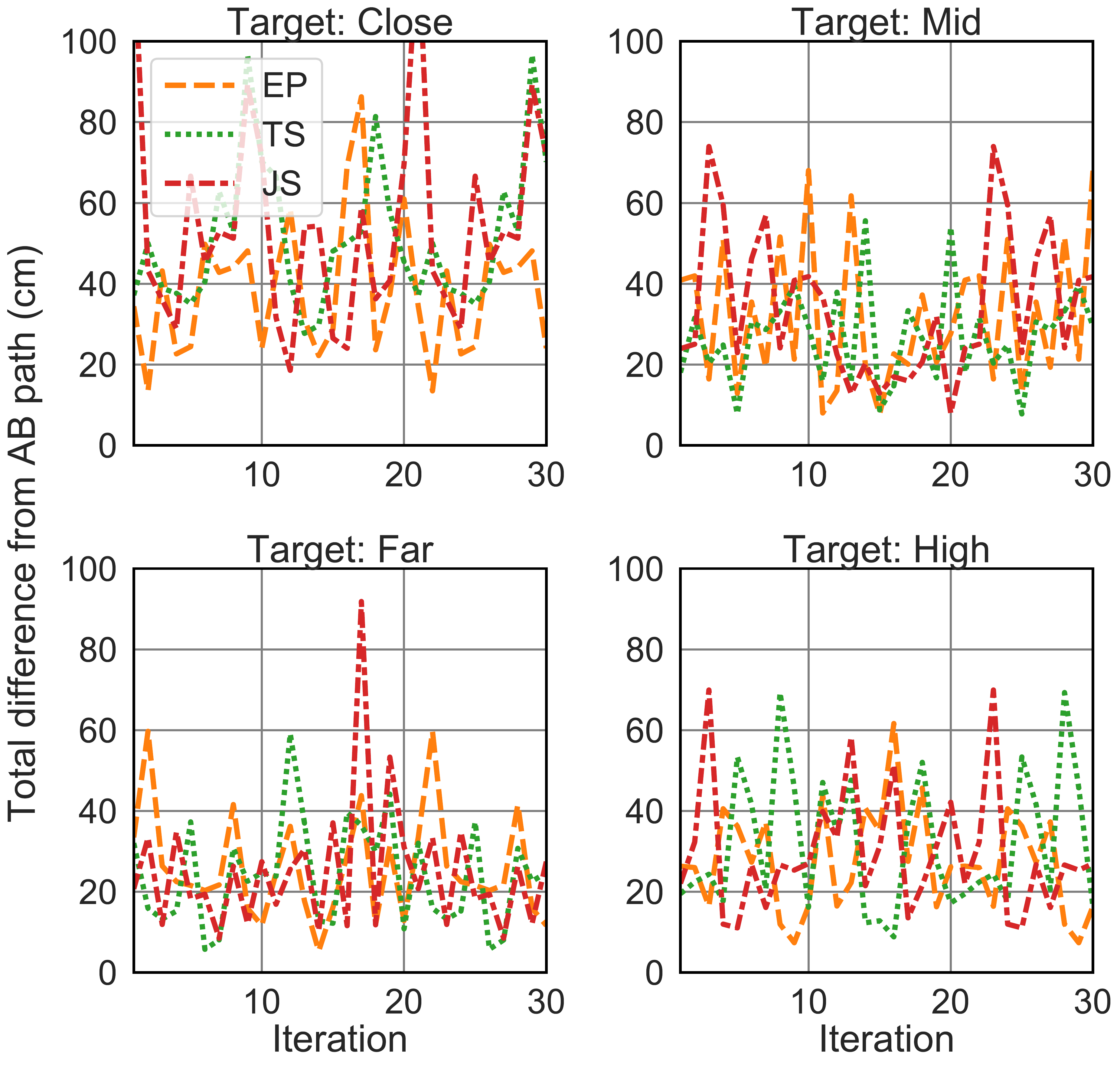}
\caption{Case study experimental results for hand path difference from AB ($\Delta_\mu$) over iterations of the task for each target.}
\label{fig:handPathDiff_prost}
\end{figure}

\begin{figure}[ht]
\centering
\includegraphics[trim={0pt 0pt 0pt 0pt}, width=0.85\columnwidth, clip]{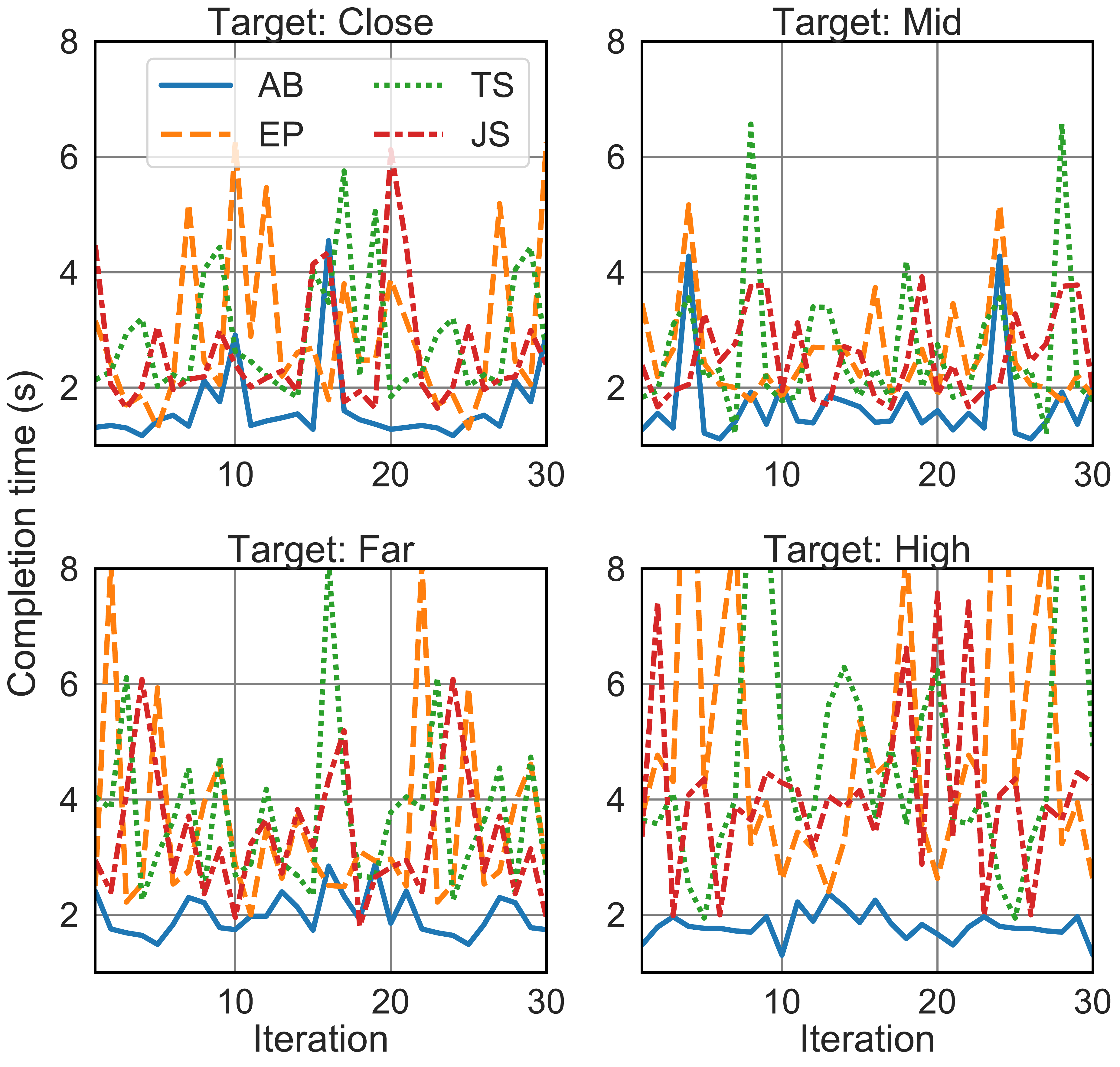}
\caption{Case study experimental results for task completion time ($t_f^{\mu}$) over iterations of the task for each target.}
\label{fig:task time_prost}
\end{figure}

\begin{figure*}[htb]
\centering
    \begin{subfigure}[t]{0.4\textwidth}
        \centering
        \includegraphics[trim={20pt 0pt 60pt 0pt}, width=\columnwidth, clip]{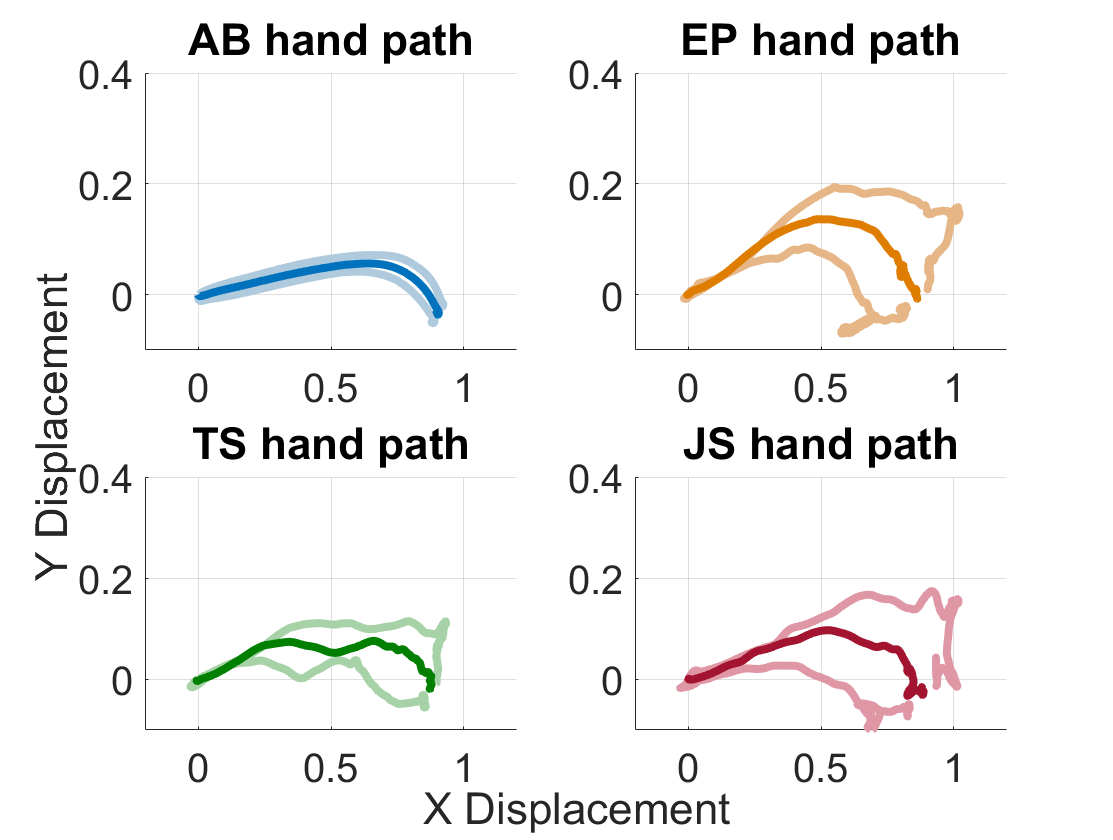}
        \caption{Mean and standard deviation hand paths.}
        \label{fig:prosHandPath}
    \end{subfigure}
    ~
    \begin{subfigure}[t]{0.4\textwidth}
        \centering
        \includegraphics[trim={20pt 0pt 50pt 0pt}, width=\columnwidth, clip]{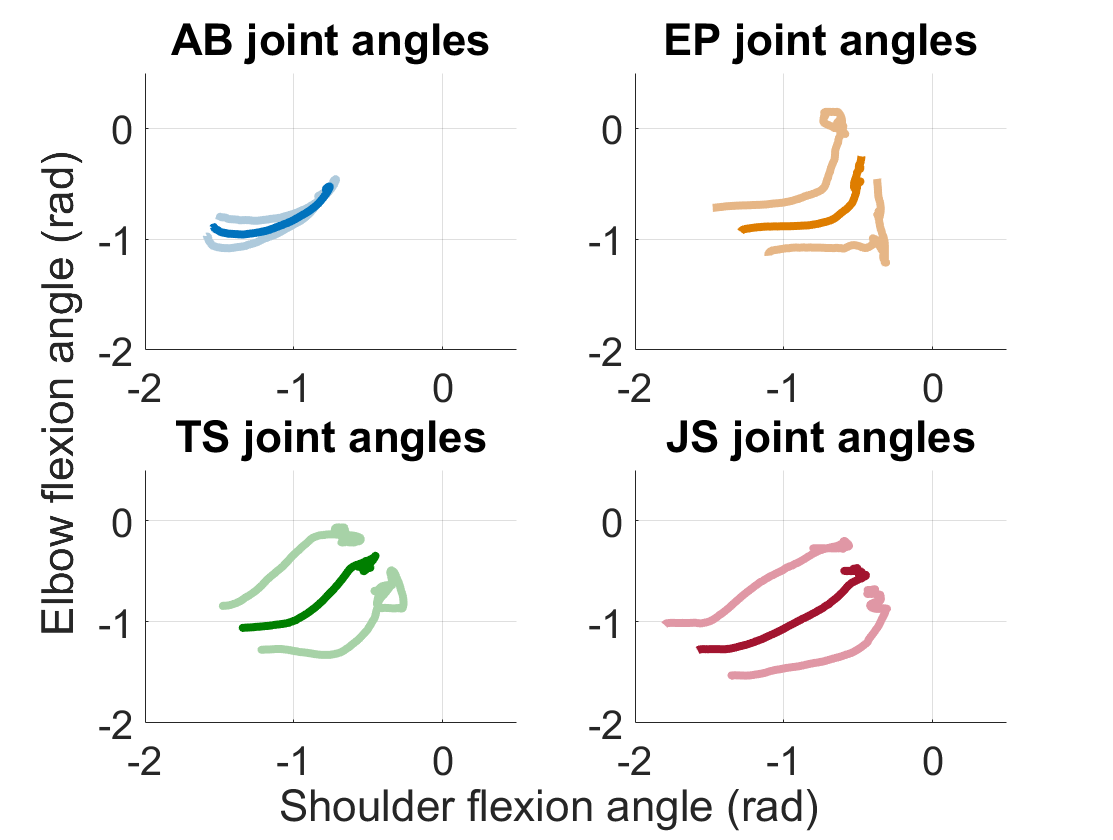}
        \caption{Mean and standard deviation joint paths.}
        \label{fig:prosJointSyn}
    \end{subfigure}
    \vspace{-2pt}
    \caption{Mean and standard deviation hand paths and joint angles of the last 10 iterations (steady-state) of the Far target for the amputee subject}
    \label{fig:prosResults}
\end{figure*}

\begin{figure*}[ht]
\centering
    %
    \begin{subfigure}[t]{0.29\textwidth}
        \centering
        \includegraphics[width=1.0\textwidth]{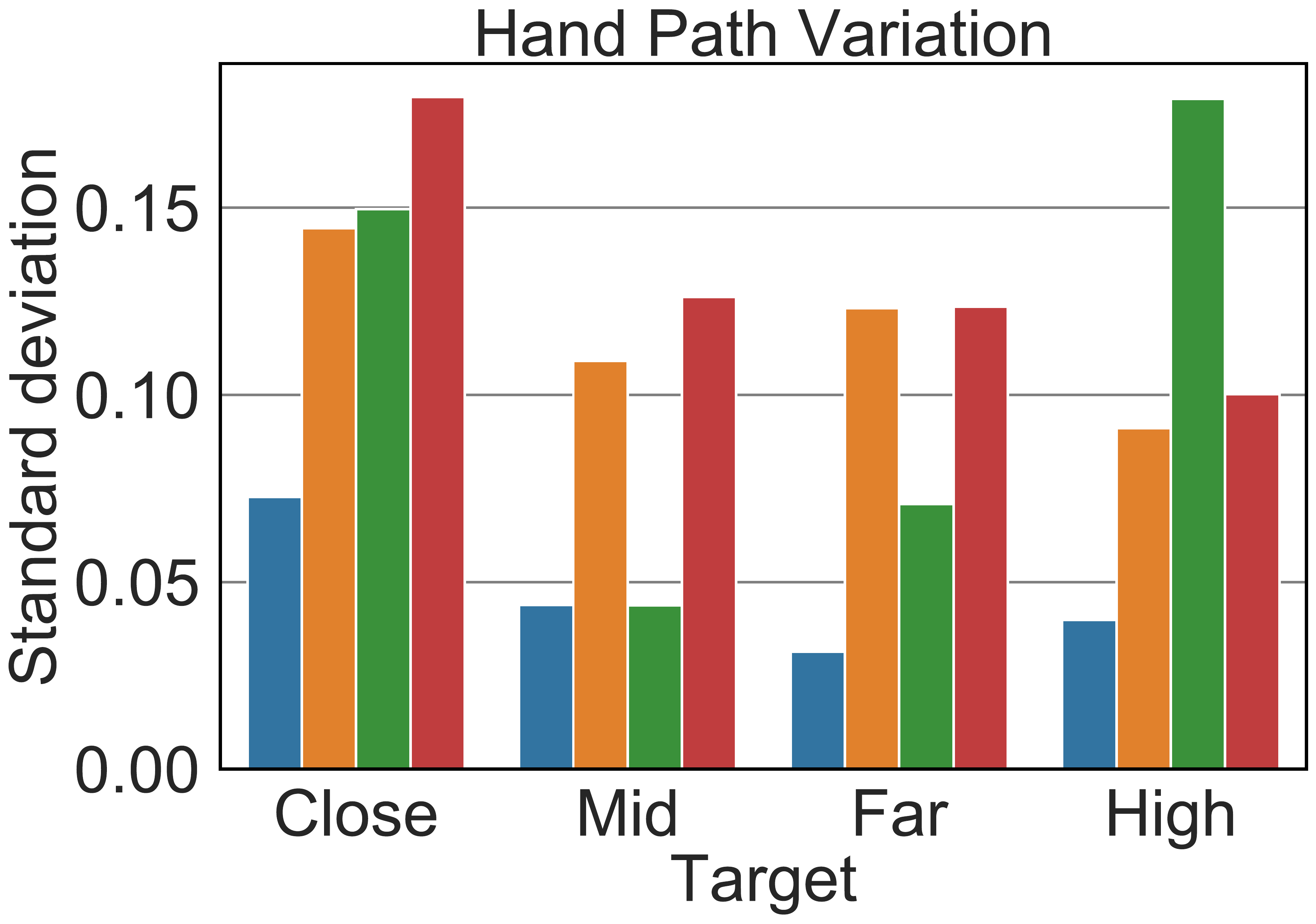}
        \caption{Amputee subject hand path variability ($\sigma_{\mu}$) over the last 10 iterations.}
        \label{fig:handVar_pros}
    \end{subfigure}
    ~
    %
    \begin{subfigure}[t]{0.29\textwidth}
        \centering
        \includegraphics[width=1.0\textwidth]{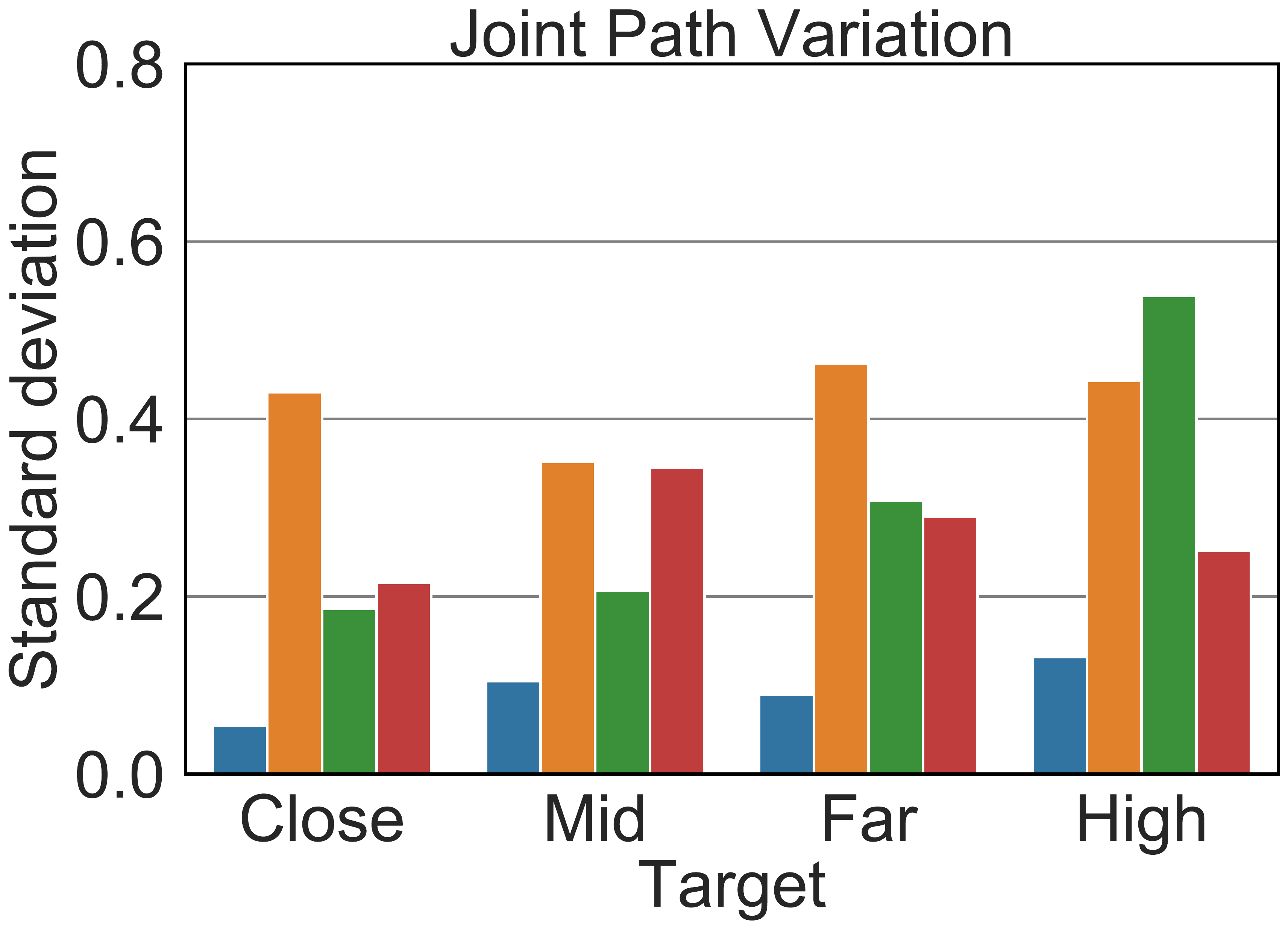}
        \caption{Amputee subject joint path variability ($\sigma_{\mu}$) over the last 10 iterations.}
        \label{fig:jointVar_pros}
    \end{subfigure}
    ~
    %
    \begin{subfigure}[t]{0.29\textwidth}
        \centering
        \includegraphics[width=1.0\textwidth]{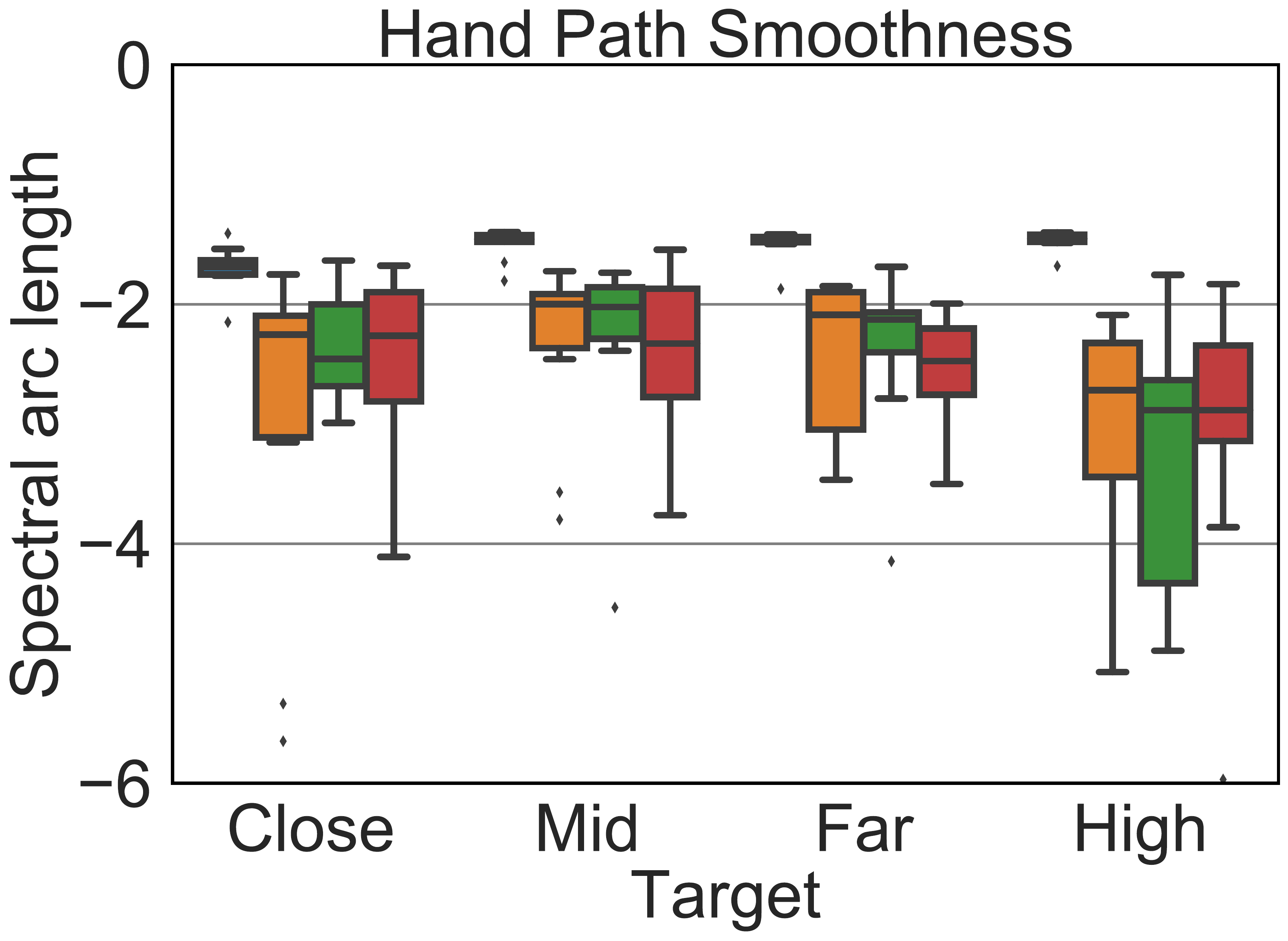}
        \caption{Amputee subject hand path smoothness ($\eta$). The less negative the smoother the path.}
        \label{fig:handSAL_pros}
    \end{subfigure}
    
    %
    \begin{subfigure}[t]{0.29\textwidth}
        \centering
        \includegraphics[width=1.0\textwidth]{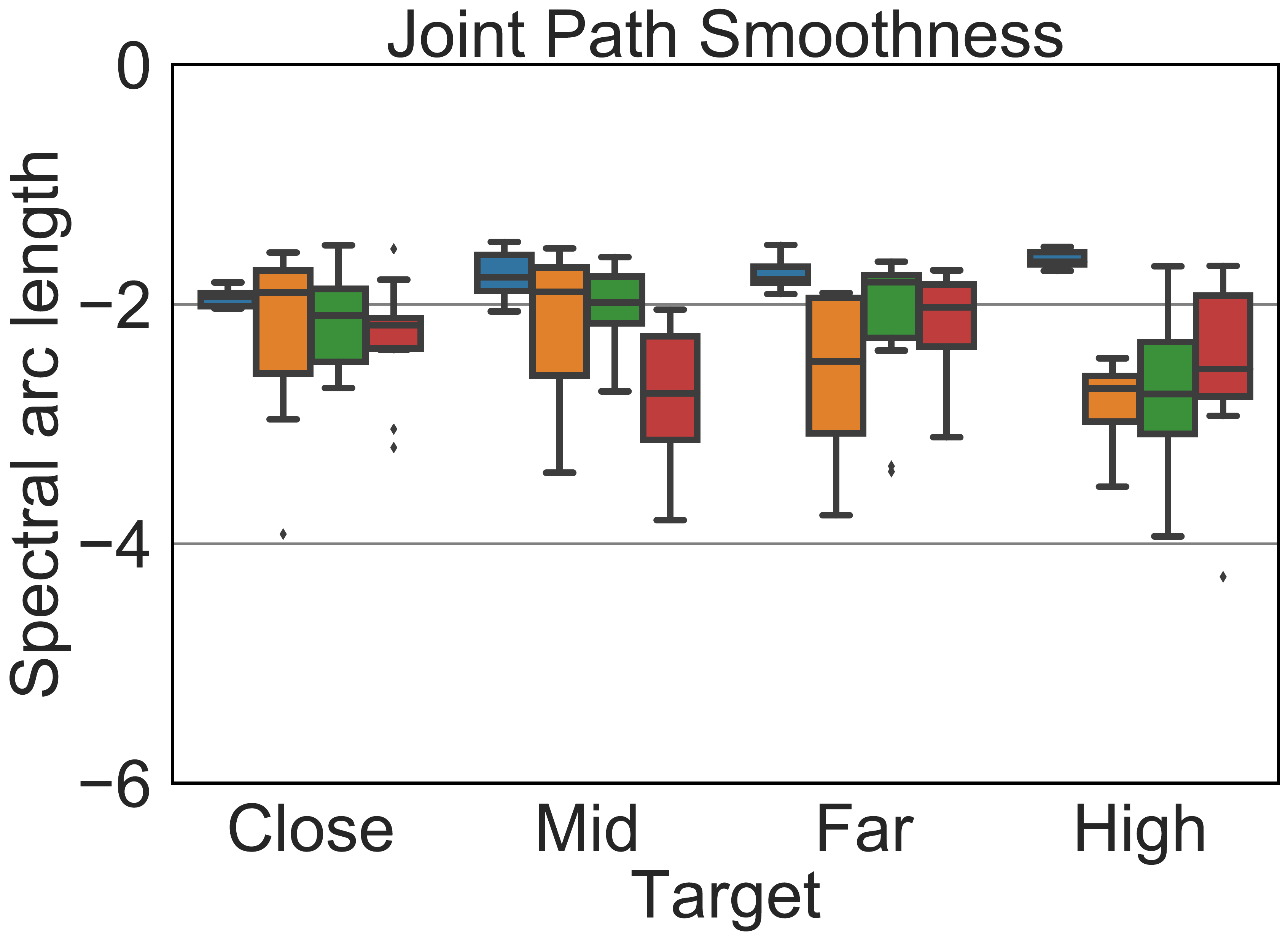}
        \caption{Amputee subject joint path smoothness ($\eta$). The less negative the smoother the path.}
        \label{fig:jointSAL_pros}
    \end{subfigure}
    ~
    %
    \begin{subfigure}[t]{0.29\textwidth}
        \includegraphics[width=1.0\textwidth]{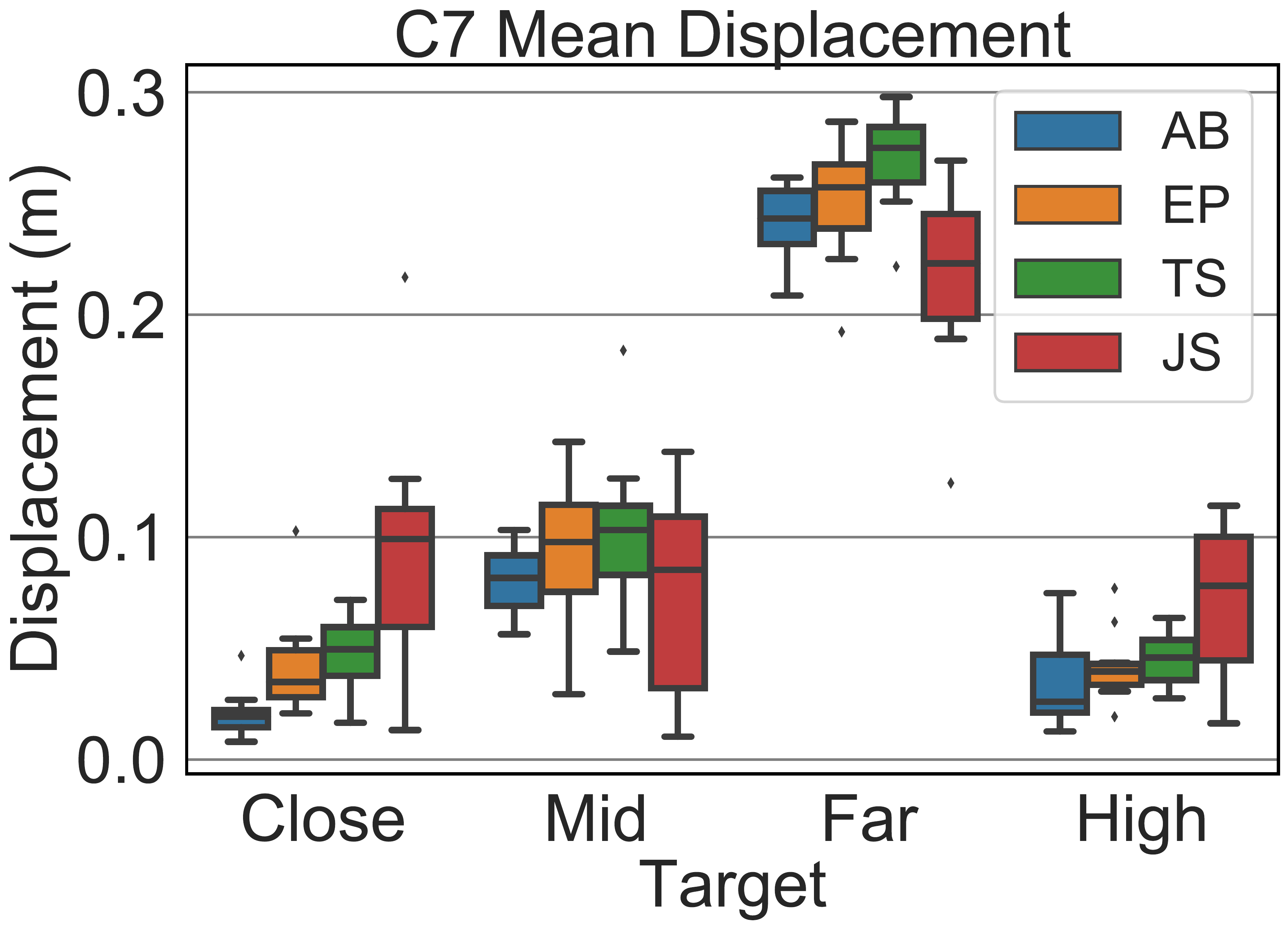}
        \caption{Amputee subject C7 displacement ($\phi^{\mu}$).}
        \label{fig:C7Displacement_pros}
    \end{subfigure}
    ~
    \begin{subfigure}[t]{0.29\textwidth}
        \includegraphics[width=1.0\textwidth]{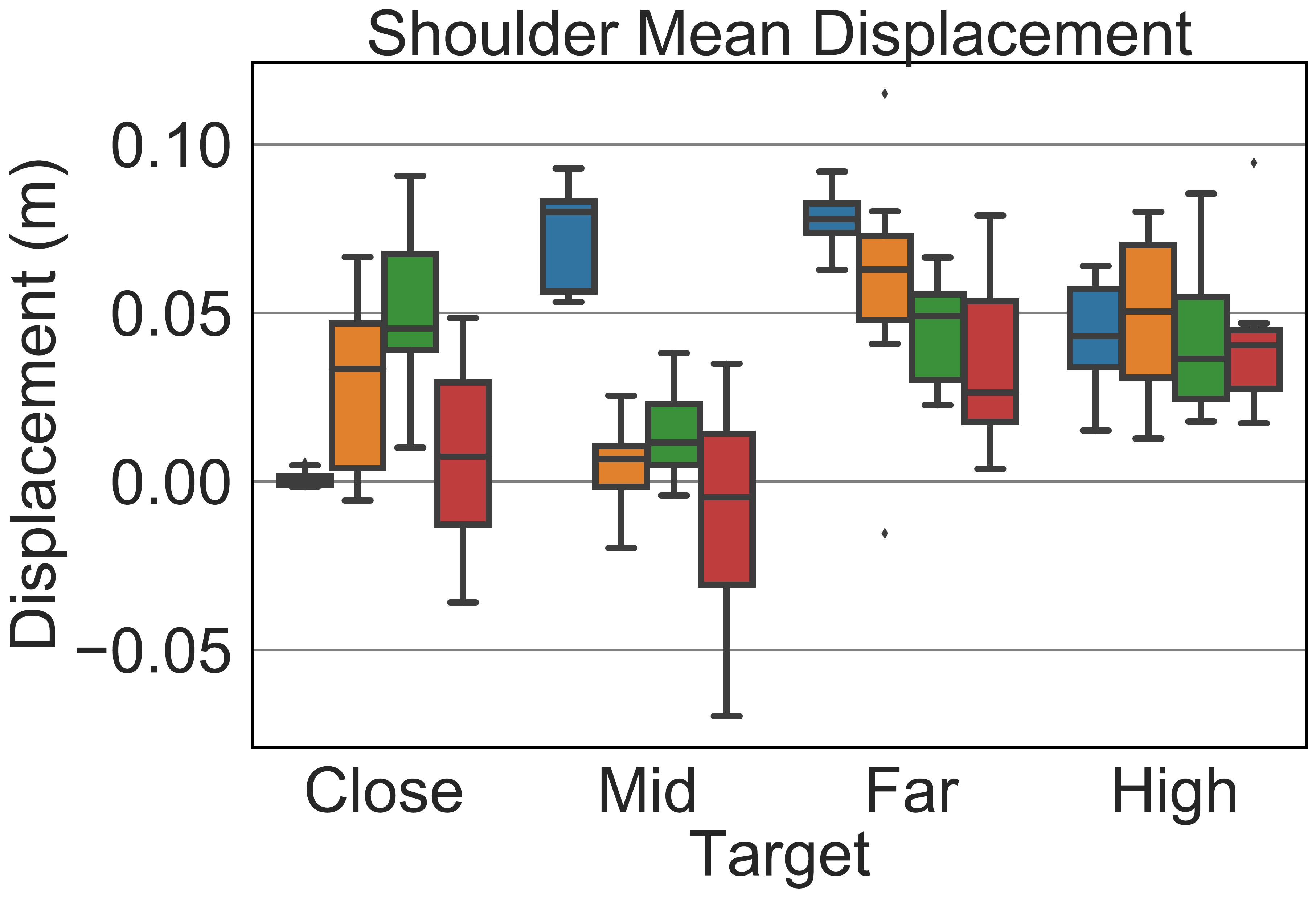}
        \caption{Amputee subject shoulder acromion displacement ($\phi^{\mu}$).}
        \label{fig:SADisplacement_pros}
    \end{subfigure}
\caption{Case study experimental results for the Quality of Motion and Motor Behaviour metrics for the last 10 iterations per target.}
\label{fig:results1_pros}
\end{figure*}

Results for the case study are presented in Figures \ref{fig:handPathDiff_prost}, \ref{fig:task time_prost}, \ref{fig:prosResults}, and \ref{fig:results1_pros}. Given the limited dataset for the case study, statistical significance cannot be obtained for the results. The results for the subject's hand path difference over iterations of the task are presented in Figure \ref{fig:handPathDiff_prost}. Moreover, the subject did not show a significant change in reaching strategy over iterations of the task, regardless of the interface modality. This may be due to reaching motion being new to the subject as he had never done this type of coordinated motion with his left arm.

The results for task completion time over iterations of the task are presented in Figure \ref{fig:task time_prost}. Similarly to hand path difference, there is no apparent improvement in time over iterations of the task. This may be a result of the short time allowed for the experiment and the novelty of the experience for the subject. As he had never used a prosthetic device nor performed the type of tasks in the experiment with his left arm, the duration of the experiment may not be sufficient to see significant improvement in his performance.

Similarly to the able-bodied subjects experimental results, the results for hand and joint paths are presented only for the Far target. Hand path results are presented in \ref{fig:prosHandPath} while joint path results are presented in \ref{fig:prosJointSyn}. For this subject, the TS case is the one that observes the lowest hand path variability (Figure \ref{fig:prosHandPath}), which is supported by the results in Figure \ref{fig:handVar_pros}, except for the High target. However, in terms of the closeness to the AB hand path, all three prosthetic cases perform similarly. On the other hand, the experimental results suggested that the JS modality achieved the lowest variability. 

Regarding the joint path, the case study results in Figure \ref{fig:prosJointSyn} are closer to the experimental results in Figure \ref{fig:repJointSyn} in terms of their shape. While the amputee subject showed smaller joint-range utilisation, similar features highlighting the interface use strategy arise. For instance, the EP strategy shows the same straight line shapes characteristic of sequential motion, while both TS and JS show curves. In the TS case, it can be seen that the amputee subject also followed the aim-and-extend strategy as seen by the horizontal line segment with lower variability. In general, both TS and JS outperform EP in terms of joint path variability as shown in Figure \ref{fig:jointVar_pros}.

Lastly, results for hand and joint path smoothness (Figures \ref{fig:handSAL_pros} and \ref{fig:jointSAL_pros}), and trunk and shoulder displacement (Figures \ref{fig:C7Displacement_pros} and \ref{fig:SADisplacement_pros}) show no significant difference between the three prosthetic cases.

These results highlight human-to-human variation, where preference to different interface modalities may arise. Moreover, the type of amputation, or in this case limb difference, may affect the subject's preference towards an interface modality. Anecdotally, the subject stated a preference towards the TS modality as he found it easier to use because of the point-and-extend strategy. This motivates further studies with subjects with different amputations and limb differences that focus on the human aspect of the human-prosthetic interface.

%% file: 8_conclusion_v2.tex
%
%
\section{Conclusion}
In this paper a realisation of a task-space synergy for motion-based human-prosthetic interfaces was presented which allows the prosthesis to be controlled in task-space through the anticipated path of the prosthetic hand. Such type of synergistic interface has the potential to be generalisable to any reaching task, is personalised through the use of a kinematic model of the user's arm, and may prove useful in complex tasks that require the coordination of multiple degrees of freedom.  An experimental study with able-bodied subjects and a case study with an amputee subject showed that the proposed task-space synergy achieves comparable performance to joint-space synergistic interfaces in terms of the closeness to able-bodied motor behaviour and task performance. These results show that the proposed task-space synergy method provides the advantage of personalisation without relying on time-consuming calibration processes or leaving it to the human to learn. Future studies will investigate the application of task-space synergies to higher complexity tasks and multi-degree-of-freedom prosthetic devices, and the case when full information on the reaching target is available.